\documentclass[a4paper,fleqn]{cas-dc}

\usepackage{soul} 
\usepackage{color, xcolor}

\usepackage[numbers]{natbib}
\usepackage{subfigure}
\usepackage{tabularx}
\usepackage{tikz}
\usepackage{pgfplots}
\usepackage[edges]{forest}
\usepackage{adjustbox}
\usepackage{supertabular}
\usepackage{caption}
\usepackage{cuted} 
\usepackage{makecell}
\usepackage{diagbox}
\usepackage[font=small,labelfont=bf,labelsep=none]{caption}
\usepackage{stfloats}

\usepackage{tabularx}

\begin{document}

\shorttitle{Artificial Intelligence in Landscape Architecture: A Survey} 
\shortauthors{Y. Xing \textit{et al.}}

\author[1]{Yue Xing}
\ead{xingyue9812@gmail.com}
\address[1]{School of Art and Design, Shaoguan University, Shaoguan 512005, China} 

\author[2]{Wensheng Gan}
\cortext[cor1]{Corresponding author}
\ead{wsgan001@gmail.com}
\address[2]{College of Cyber Security, Jinan University, Guangzhou 510632, China} 
\cormark[1]

\author[1]{Qidi Chen}
\ead{1310428439@qq.com}

\title [mode = title]{Artificial Intelligence in Landscape Architecture: A Survey}

\begin{abstract}
   The development history of landscape architecture (LA) reflects the human pursuit of environmental beautification and ecological balance. With the advancement of artificial intelligence (AI) technologies that simulate and extend human intelligence, immense opportunities have been provided for LA, offering scientific and technological support throughout the entire workflow. In this article, we comprehensively review the applications of AI technology in the field of LA. First, we introduce the many potential benefits that AI brings to the design, planning, and management aspects of LA. Secondly, we discuss how AI can assist the LA field in solving its current development problems, including urbanization, environmental degradation and ecological decline, irrational planning, insufficient management and maintenance, and lack of public participation. Furthermore, we summarize the key technologies and practical cases of applying AI in the LA domain, from design assistance to intelligent management, all of which provide innovative solutions for the planning, design, and maintenance of LA. Finally, we look ahead to the problems and opportunities in LA, emphasizing the need to combine human expertise and judgment for rational decision-making. This article provides both theoretical and practical guidance for LA designers, researchers, and technology developers. The successful integration of AI technology into LA holds great promise for enhancing the field's capabilities and achieving more sustainable, efficient, and user-friendly outcomes.
\end{abstract}


\maketitle

\section{Introduction}

Landscape architecture (LA) \cite{newton1971design,francis2001case} is a beautiful place where humans and nature are integrated, providing valuable green spaces for cities and offering leisure, entertainment, and cultural experiences for people. Specifically, LA refers to the creation of landscape spaces with aesthetic value and human significance through the arrangement of the natural environment and artistic conception, meeting people's needs for leisure, recreation, and appreciation of nature. It is an interdisciplinary subject involving landscape design, garden planning, plant configuration, water landscape construction, and other aspects. With the rapid development of urbanization and increasing environmental pressures, scholars have begun to explore the relevant applications of artificial intelligence (AI) \cite{nilsson1982principles,zhang2021study} in LA.

AI can accurately and scientifically identify and analyze the preliminary conditions of a project, establish design logic, evaluate design results and network computing, output actual design results, perform intelligent calculations based on parameter inputs, and generate optimal solutions \cite{giones2017digital}. In recent years, AI technology has carried out a lot of application research in related fields dominated by architecture \cite{pena2021artificial}, planning \cite{son2023algorithmic}, and landscape \cite{cantrell2021artificial}, bringing new development opportunities for LA and providing powerful tools and resources for designers, planners, and managers to plan better, design and manage landscape spaces. AI technology also can provide scientific support for objectively understanding the development laws of LA \cite{newton1971design}.

Firstly, AI plays an important role in LA's planning and design process. The traditional design process \cite{filor1994nature} requires a lot of time and manpower, while AI can accelerate the design process and provide diverse design options through machine learning and data analysis. For example, using generative adversarial network (GAN) \cite{goodfellow2014generative,creswell2018generative} can generate realistic virtual landscape images, creating an interactive, virtual-real landscape environment \cite{urban2021designing} to help designers visualize concepts and solutions. In addition, AI can also perform shape optimization and automatic layout generation \cite{shi2023Iintelligent}, providing inspiration and creative support.

Secondly, AI can improve the efficiency of LA management and maintenance. AI, through intelligent monitoring and forecasting, helps managers identify management and maintenance issues in a timely manner and take appropriate measures. For example, the smart irrigation and maintenance system \cite{goap2018iot} uses sensors and data analysis to monitor the water and health status of plants, and intelligently adjusts the irrigation amount and frequency to achieve water saving and optimize plant growth. In addition, AI can also identify and monitor pests and diseases \cite{chen2020aiot}, providing early warning and precise prevention and control measures. Through ecosystem simulation and assessment \cite{bousquet2004multi}, AI can predict indicators such as vegetation growth, water resource utilization, and biodiversity, helping designers and managers understand the impact of different design schemes on the ecosystem and optimize design and management strategies. AI can solve problems such as environmental destruction and ecological degradation through data analysis and model prediction \cite{papadimitriou2012artificial}, promoting sustainable development.

Finally, in addition to design, construction, and management, AI can also improve visitor experience in self-guided tour systems and interactive experience applications during the tourism process. Using machine learning (ML) \cite{alpaydin2020introduction,jordan2015machine} and natural language processing (NLP) technologies \cite{collobert2011natural}, it can provide personalized tours and recommendations \cite{vansteenwegen2011city} for visitors based on their interests and preferences, recommending attractions and activities. Furthermore, the use of computer vision (CV) \cite{szeliski2022computer} and metaverse \cite{wang2022survey} technologies can achieve attraction identification and augmented reality experiences \cite{scholz2016augmented}, providing visitors with a richer, more interactive, and immersive experience.

\begin{figure}[ht]
    \centering
    \includegraphics[width=0.98\linewidth]{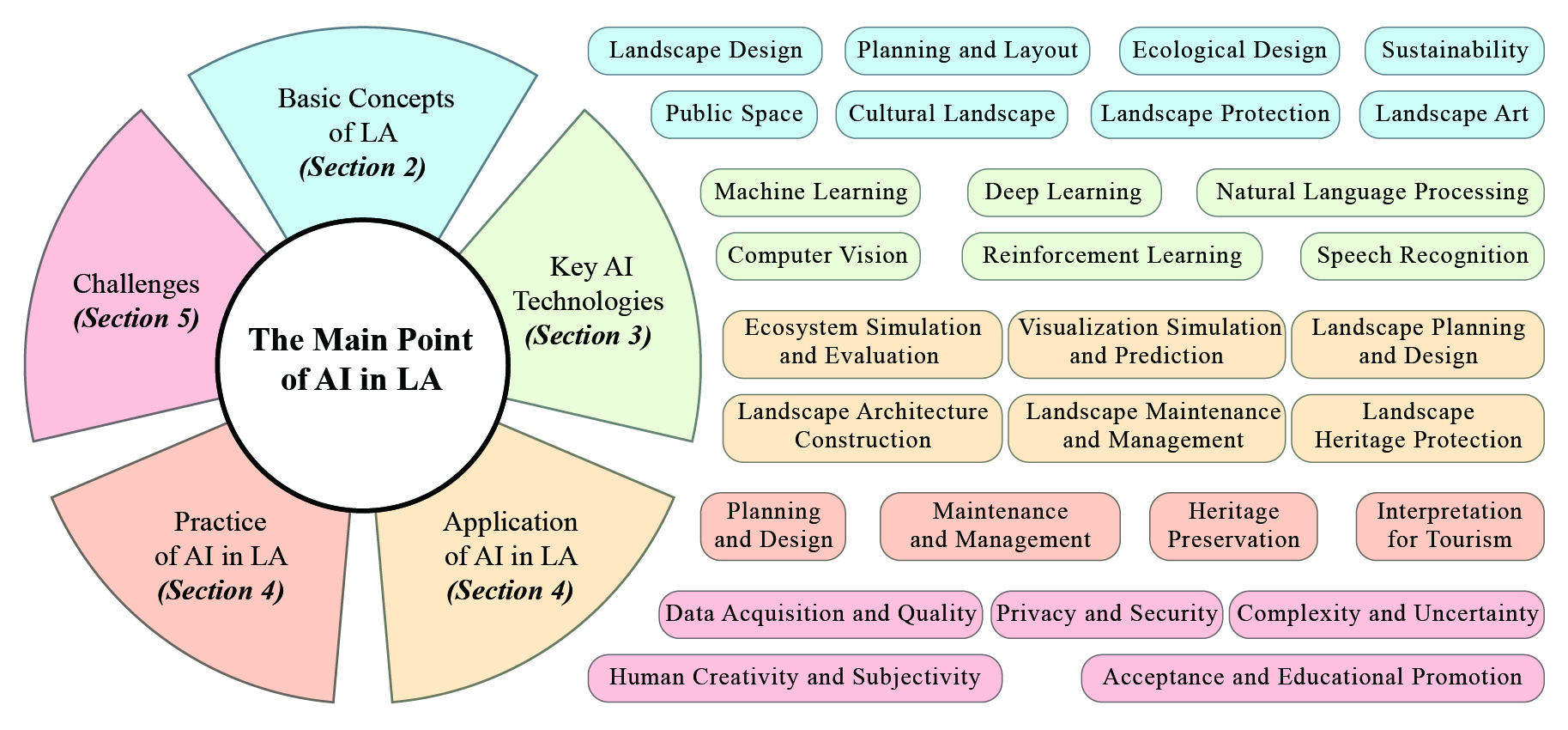}
    \caption{The main points of AI in LA in this paper.}
    \label{fig:main point}
\end{figure}

However, the application of AI in LA also faces some important challenges. For example, in terms of data acquisition and quality, AI requires a large amount of high-quality data for training and learning \cite{Roh202survey}, but in LA, the acquisition and organization of these data may have a certain complexity and cost. In addition, protecting user privacy and data security \cite{liu2020privacy} is also an important issue, which needs to be properly handled and protected in AI applications. Furthermore, the application of AI in LA needs to be combined with human professional knowledge and creativity, rather than replacing the role of humans. Human aesthetics, emotions, and judgment are key elements in the design and management of LA, and AI should be an auxiliary tool and resource. Finally, regarding the promotion and popularization of AI technology, although AI has great potential in LA, its application is still relatively new and immature. It is necessary to strengthen related research and practice, cultivate professional talents, and promote cooperation and exchange among academia, industry, and government to promote the widespread applications of AI technologies in LA. The main points of AI in LA  are shown in Figure \ref{fig:main point}.

In summary, the AI applications in the field of LA provide new possibilities for solving the problems and challenges currently faced. By accelerating the design process, optimizing management and maintenance, enhancing ecological friendliness, and improving visitor experience, AI can bring greater benefits and sustainable development to LA. However, we need to deeply study and explore this field, overcome the technical and implementation difficulties, and realize the maximum potential and value of AI in LA. The main contributions of this paper include:

\begin{itemize}
    \item First, it introduces the basic concept and characteristics of AI, and the application of AI in LA is gradually increasing, bringing many potential benefits to design, planning, and management.

    \item It specifically discusses how AI can assist the field of LA in solving the current development problems, including urbanization, environmental destruction, ecological degradation, unreasonable planning, insufficient management and maintenance, and lack of public participation.

    \item It summarizes in detail the key technologies and related practical cases of using AI for LA design, and analyzes the importance of applying AI to LA design. It also demonstrates the specific applications of AI in LA through case studies, from design assistance to intelligent management, which provides new possibilities and innovative solutions for the planning, design, and maintenance of LA.

    \item  Finally, it highlights some existing problems and opportunities. The potential of AI in LA is huge, but it still needs to be combined with human professional knowledge and judgment to make reasonable decisions. The role of professionals in the design, planning, and management process is still crucial.
\end{itemize}

\begin{figure}[ht]
    \centering
    \includegraphics[width=0.95\linewidth]{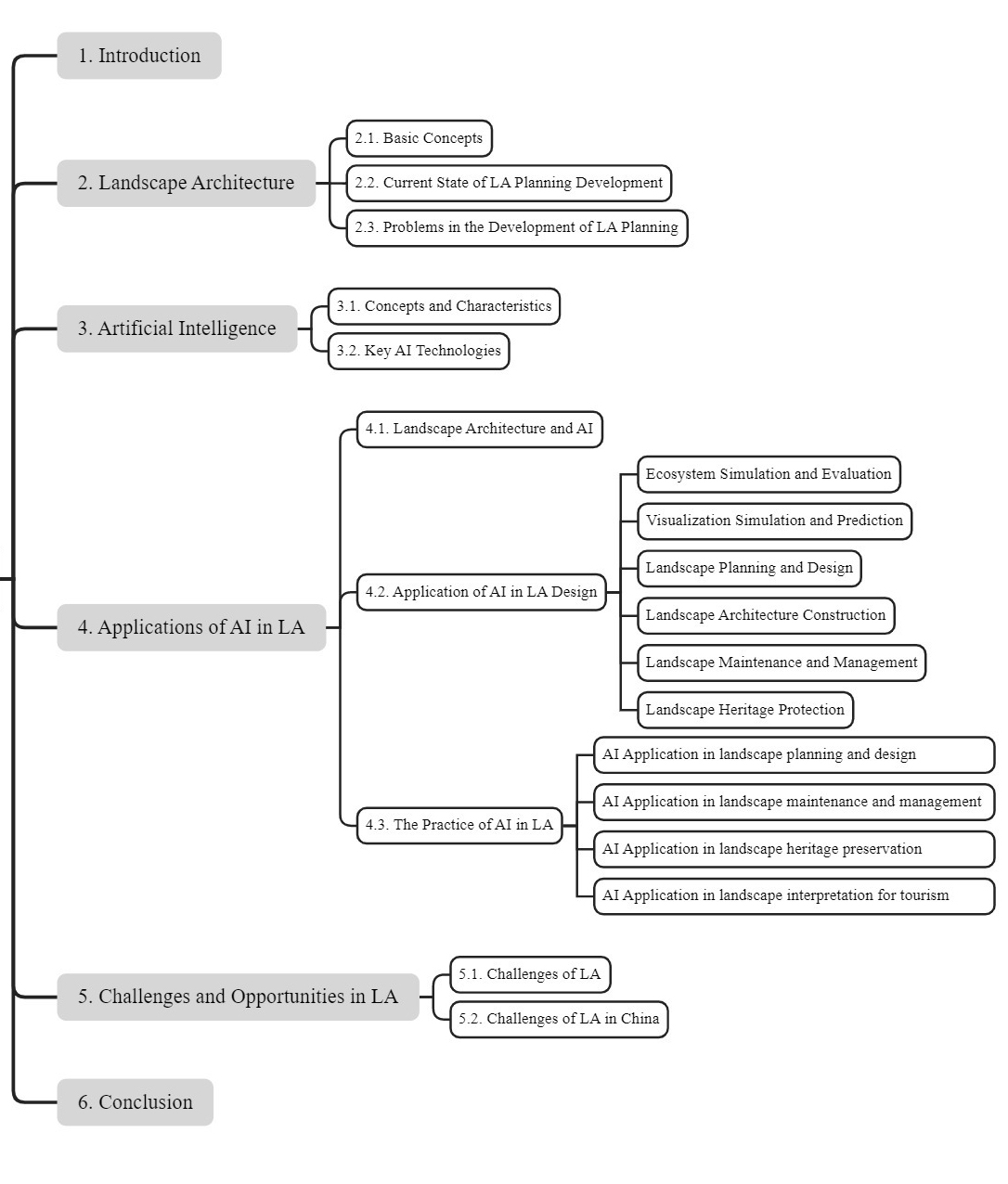}
    \caption{The outline of this paper.}
    \label{fig:outline}
\end{figure}

The organization of this paper is as follows: Section \ref{sec:landscape} gives background about the concepts, current state, and problems content of LA planning. Section \ref{sec:AI} provides key concepts and characteristics of AI, as well as key AI technologies. Section \ref{sec:applications} focuses on different applications of AI in LA. Some challenges and opportunities about AI in LA are respectively discussed in section \ref{sec:challenges}. Finally, Section \ref{sec:conclusion} presents the conclusion. Details of the outline of this paper are presented in Figure \ref{fig:outline}.

\section{Landscape Architecture}\label{sec:landscape}

The creation process of LA needs to consider factors such as spatial layout, landscape elements (e.g., plants, water bodies, stone materials, etc.), topography, cultural history, and environmental ecology, and integrate people's needs and aesthetic preferences in the design and planning \cite{newton1971design,francis2001case}. Landscape architecture can include parks, urban greenspaces, squares, courtyards, scenic areas, and other landscape environments. Landscape designers and planners utilize various design techniques and technologies, combined with people's needs and site characteristics, to apply landscape elements like topography, water bodies, plants, structures, and pavement, to create diverse spatial forms and landscape features. Besides, the design and planning of LA need to consider environmental protection, ecological balance, and sustainable development, to achieve harmonious coexistence between humans and nature.

\subsection{Basic Concepts}
% 1.1 Relevant Concepts in LA
Landscape architecture (LA) is an art and science field that focuses on the integration of human and natural environments, involving the planning, design, construction, and management of various public and private landscape spaces, to create beautiful, functional, and sustainable environments. The following are some important concepts related to LA:

\textbf{Landscape design} \cite{motloch2000introduction,levinthal1999landscape} is a core aspect of LA, involving the creation and improvement of the aesthetic, functional, and environmental quality of landscape spaces. Landscape designers combine natural elements (such as plants, water bodies, and topography) and man-made elements (such as roads, buildings, and sculptures) through artistic and scientific means to create landscapes that are adapted to human life and activities.

\textbf{Planning and layout} \cite{marsh2005landscape} refers to the overall planning and organization of LA spaces according to specific goals and requirements. Planning includes determining land uses, functional zoning, and spatial layout, to ensure the rationality, sustainability, and social benefits of the LA.

\textbf{Ecological design} \cite{van2013ecological} is a design approach that emphasizes the protection and restoration of natural ecosystems. It stresses the integration with the natural environment, by selecting plant species adapted to the local climate, soil, and vegetation, to restore ecological functions, provide habitats for wildlife, and minimize negative environmental impacts as much as possible.

\textbf{Sustainability} \cite{leitao2002applying,wu2013landscape} refers to the comprehensive consideration of social, economic, and environmental factors in the design and management of LA, to meet current needs without compromising the ability of future generations. Sustainability includes efforts in the rational use of resources, energy conservation, water resource management, and reduction of waste and pollution.

\textbf{Public space} \cite{carr1992public} refers to the landscape spaces in cities and communities that are open to the public, such as parks, squares, streets, and landscape corridors. Public spaces provide places for people's leisure, entertainment, cultural, and social activities, fostering community cohesion and social interaction.

\textbf{Cultural landscape} \cite{taylor2011cultural} is a landscape that combines natural and human cultural elements. It includes areas with historical, artistic, religious, or cultural significance, such as historic sites, historical parks, and cultural heritage sites. Cultural landscapes reflect the interaction and evolution process between humans and the natural environment.

\textbf{Landscape protection} \cite{beresford2000protected} refers to the protection and maintenance of landscape resources with natural, cultural, and aesthetic value. It includes the protection of natural ecosystems, cultural heritage, historic buildings, and landscape elements, to ensure sustainability and inheritance.

\textbf{Landscape art} \cite{milani2009art} is the creative integration of natural and man-made elements to create landscapes with aesthetic value and emotional appeal. Landscape art involves artistic expression in aspects, such as landscape composition, color application, material selection, and light and shadow effects.

These concepts collectively form the foundation of the LA field and play an important guiding role in the design and management of landscape spaces. The goal of LA is to create pleasant, sustainable, and harmonious landscape environments that provide people with places for leisure, entertainment, social interaction, and cultural exchange.

\subsection{Current State of LA Planning Development}
LA planning has seen significant development and promotion over the past decades. In general, the field exhibits the following development status:

\textbf{(1) Urban park construction} \cite{cranz2004defining}: The construction of urban parks is thriving. Many cities are committed to increasing green space and improving the ecological environment to provide places for people's leisure and recreation. Large cities like New York, Beijing, and Guangzhou have built comprehensive parks, themed parks, and water parks.

\textbf{(2) Protection of historical and cultural heritage sites} \cite{carter1997balancing}: There are abundant historical and cultural heritage sites, including ancient Roman ruins, ancient city walls, royal gardens, and temples. To protect and pass down these sites, large-scale cultural heritage protection and restoration work has been carried out, and relevant planning and management measures have been formulated.

\textbf{(3) Promotion of ecological landscape} \cite{turner1989landscape}: Ecological landscape emphasizes biodiversity conservation, water resource management, and sustainable development. Many cities have built ecological parks and ecological corridors, restoring and protecting natural landscapes such as wetlands, rivers, and forests.

\textbf{(4) Planning of urban-rural fringe areas} \cite{gallent2007spatial,li2021understanding}: These areas have both urban and rural characteristics and require rational planning and management. The rural revitalization strategy has been promoted, creating a livable rural environment through planning and design, and improving the quality of life for farmers.

\textbf{(5) Tourist attraction planning} \cite{inskeep1987environmental}: Planning of tourist attraction has also made significant progress. For famous tourist attractions, such as the Forbidden City, the Great Wall, and Mount Huangshan, planning and managing these scenic areas is crucial for providing a good sightseeing experience. Domestic and foreign guidelines for tourism destination planning and management have been developed, strengthening the protection, management, and sustainable development of scenic areas.

In summary, LA planning is constantly evolving, focusing on ecological environment protection, historical and cultural heritage preservation, and urban greening construction. The Chinese government and urban management departments have taken a series of measures in planning, design, and management to improve the quality of life and the sustainable development of the urban environment.

\subsection{Problems in Development of LA Planning}

The current LA planning field faces some problems, especially in some countries like China, including but not limited to the following main issues:

\begin{itemize}
    \item \textbf{Urban expansion pressure}: With the acceleration of urbanization, the urban area is constantly expanding, and the demand for LA space is also increasing. However, urban development is often driven by economic interests, leading to the compression and destruction of LA space. Urban expansion pressure poses certain challenges to the protection and development of LA.

    \item \textbf{Environmental damage and ecological degradation}: Some LA areas have experienced environmental damage and ecological degradation due to over-development, over-exploitation, and improper management. For example, excessive land reclamation, over-exploitation of water resources, and vegetation destruction have all had negative impacts on the ecosystem, threatening the living environment of plants and animals.

    \item \textbf{Unreasonable planning and design}: Some LA planning and design have problems, including lack of overall planning, unreasonable layout, lack of innovation, and lack of sustainability. Some LA projects focus too much on surface effects while neglecting functionality and environmental adaptability, leading to unreasonable space utilization and imbalance in the ecological system.

    \item \textbf{Insufficient management and maintenance:} Some LA projects face challenges in long-term management and maintenance. Management lacks effective mechanisms and professional teams, leading to facility damage and neglect. Lack of long-term funding investment and reasonable management plans will lead to management problems in parks and scenic areas.

    \item \textbf{Low public participation}: In some LA projects, public participation is low, and the demands and opinions of the public are not adequately considered. This may lead to public dissatisfaction and resistance to the garden projects, affecting their sustainable development and social acceptance.
\end{itemize}

These problems require the joint efforts of the government, planning and design agencies, professional teams, and the public to solve. By strengthening sustainable planning and design, promoting ecological protection and restoration, enhancing management and maintenance, and encouraging public participation, the quality and sustainable development level of LA can be improved.

\section{Artificial Intelligence} \label{sec:AI}
\subsection{Concepts and Characteristics of AI}

The concept of artificial intelligence (AI) was first proposed by John McCarthy, at the Dartmouth conference in 1956 \cite{nilsson1982principles}. It simulates and expands human intelligence, with the basic goal of enabling computers to simulate and perform tasks that usually require human intelligence \cite{guilford1967nature}, such as perception, learning, understanding, reasoning, language understanding, and generation \cite{salomon1991partners}. The main research areas of AI technology include traditional machine learning, deep learning, computer vision, robotics, knowledge engineering, etc. Based on Nilsson's perspective in 1971 \cite{nilsson1971problem}, AI refers to high-performance and complex machines that perform cognitive functions usually associated with human intelligence (such as learning, interaction, and problem-solving). Bartneck mentioned that AI can handle learning, reasoning, and problem-solving \cite{bartneck2009measurement}. With the development of intelligence, AI assistants can perform various complex tasks by learning information from users and the environment, responding to environmental changes, analyzing user preferences, and executing tasks based on user preferences \cite{hu2021can}. According to the attributes, functions, and types of LA problems that can be solved, the AI technologies currently applied in the field of LA can be mainly divided into intelligent stochastic optimization, artificial life, and machine learning.

\subsection{Key AI Technologies}

There are many key technologies of AI related to LA, covering multiple aspects, mainly including:

\textbf{Machine learning (ML)} \cite{jordan2015machine} is one of the core technologies in the field of AI, which allows computers to automatically learn patterns and rules from data to perform classification \cite{bost2014machine}, prediction \cite{mair2000investigation}, and decision-making \cite{tejeda2022ai,gan2023anomaly}. Machine learning technologies include supervised learning \cite{caruana2006empirical}, unsupervised learning \cite{dy2004feature}, reinforcement learning \cite{sutton2018reinforcement}, and others.

\textbf{Deep learning (DL)} \cite{lecun2015deep} is a new research direction of machine learning, which uses neural network models to process and analyze large amounts of data. Deep learning has achieved significant results and applications in fields such as image recognition \cite{he2016deep}, speech recognition \cite{abdel2014convolutional}, and natural language processing \cite{otter2020survey}.

\textbf{Natural language processing (NLP)} \cite{collobert2011natural} involves understanding and processing natural language. NLP technologies include part-of-speech tagging, semantic analysis \cite{gabrilovich2007computing}, sentiment analysis \cite{nasukawa2003sentiment}, and machine translation \cite{koehn2009statistical}.

\textbf{Computer vision (CV)} \cite{szegedy2016rethinking} is an important branch of AI, which involves computer understanding and processing of images and videos. CV technologies include image recognition, object detection, and image segmentation \cite{minaee2021image}.

\textbf{Reinforcement learning (RL)} \cite{sutton2018reinforcement} is a new research direction of machine learning. It optimizes strategies through trial and error and solves the problem of intelligent agents learning strategies through interaction with the environment to maximize rewards or specific goals. RL has many applications such as autonomous driving and robot control.

\textbf{Speech recognition} \cite{graves2013speech} is an important technology in the field of AI, which involves converting speech into text. Currently, speech recognition has a wide range of applications in intelligent assistants, autonomous driving, and customer service.

In addition, the development of AI also depends on the progress of hardware accelerators, such as graphics processing units (GPUs) \cite{talib2021systematic}, tensor processing units (TPUs) \cite{wang2020benchmarking}, and neural processing units (NPUs) \cite{tan2021efficient}. These specialized hardware accelerators can provide varying degrees of acceleration and optimization for different AI tasks.

\section{Applications of AI in LA} \label{sec:applications}
\subsection{Landscape Architecture and AI}

\textbf{Why LA needs AI technology}? Digital technologies have been widely applied in landscape planning, design, construction, and maintenance management. The contemporary era calls for the integration of quantitative and qualitative research in the design discipline and scientific technology \cite{bryman2006integrating}, showcasing the further fusion of design and science. 

From the overall landscape architecture workflow, (1) AI can provide inspiration support and define design problems with constraints \cite{lin2015mining,lee2024when}, assisting landscape designers in generating ideas and optimizing design schemes to improve design efficiency \cite{chen2024enhancing}. By analyzing vast amounts of landscape data, images, and geographic information, AI can provide design inspiration and automatically generate landscape layouts, which helps accelerate the design process, reduce tedious manual operations, and offer more creative possibilities for designers. (2) AI can provide data analysis and decision-making support \cite{gan2020huopm,gan2021fast}. Landscape design and management involve a large amount of data, including geographic information, climate data, soil quality, etc. AI can help interpret and utilize these data through data analysis and modeling, providing decision-making support \cite{ibrahim2023expatriates}. It can predict and evaluate the effects of different design schemes, optimize resource utilization and environmental impact, and provide scientific evidence and recommendations. (3) AI can provide ecological simulation \cite{zhai2020simulating} and sustainability assessment \cite{nishant2020artificial}. Landscape design and planning need to consider the protection of ecosystems and sustainable development. AI can simulate and evaluate the impact of different design schemes on the ecosystem, including water resource utilization, plant selection, and biodiversity protection. It can help optimize design schemes to ensure ecological friendliness and sustainability. (4) AI can provide automated maintenance and management \cite{talaviya2020implementation}. AI can be applied to automated maintenance and management systems, helping to monitor the growth status of plants, pest and disease conditions, and water demand, providing precise maintenance recommendations and reducing labor and time investment. It can optimize watering and fertilization plans to improve management efficiency and resource utilization. (5) AI can provide intelligent guidance and experiences \cite{gretzel2015smart,chen2024data}. Through intelligent guidance systems, AI can recommend personalized tour routes and activities based on visitors' interests and locations. It can provide information about scenic spots, interpretations, and interactive experiences, enhancing visitor engagement and satisfaction. 

From the perspective of future trends, the integration of LA and AI technology is a strategic choice for future development, keeping pace with or even leading the industry. In modern society, with ever-changing urban construction, the creation of a beautiful and livable living environment requires the assistance of technology. The current urgent need for urban development is to transform towards smart cities \cite{chourabi2012understanding} based on a new understanding of data \cite{hashem2016role}, revealing the new trends in urban construction and development. As AI technology advances rapidly, the progress of science and technology has also driven the development of various fields. The rapid development of AI technology has become an unavoidable issue in the process of human scientific and technological development, profoundly impacting people's lives and gradually influencing other fields such as environmental design. As early as the 1970s, there were studies on the application of AI for architecture \cite{eastman1974outline}; in the following decades, AI has been explored in various research directions of LA. With the enhancement of research depth and breadth, the transformation of LA is inseparable from the integration with AI technology, and it is gradually developing towards intelligence.

\begin{figure}[ht]
    \centering
    \includegraphics[width=1.02\linewidth]{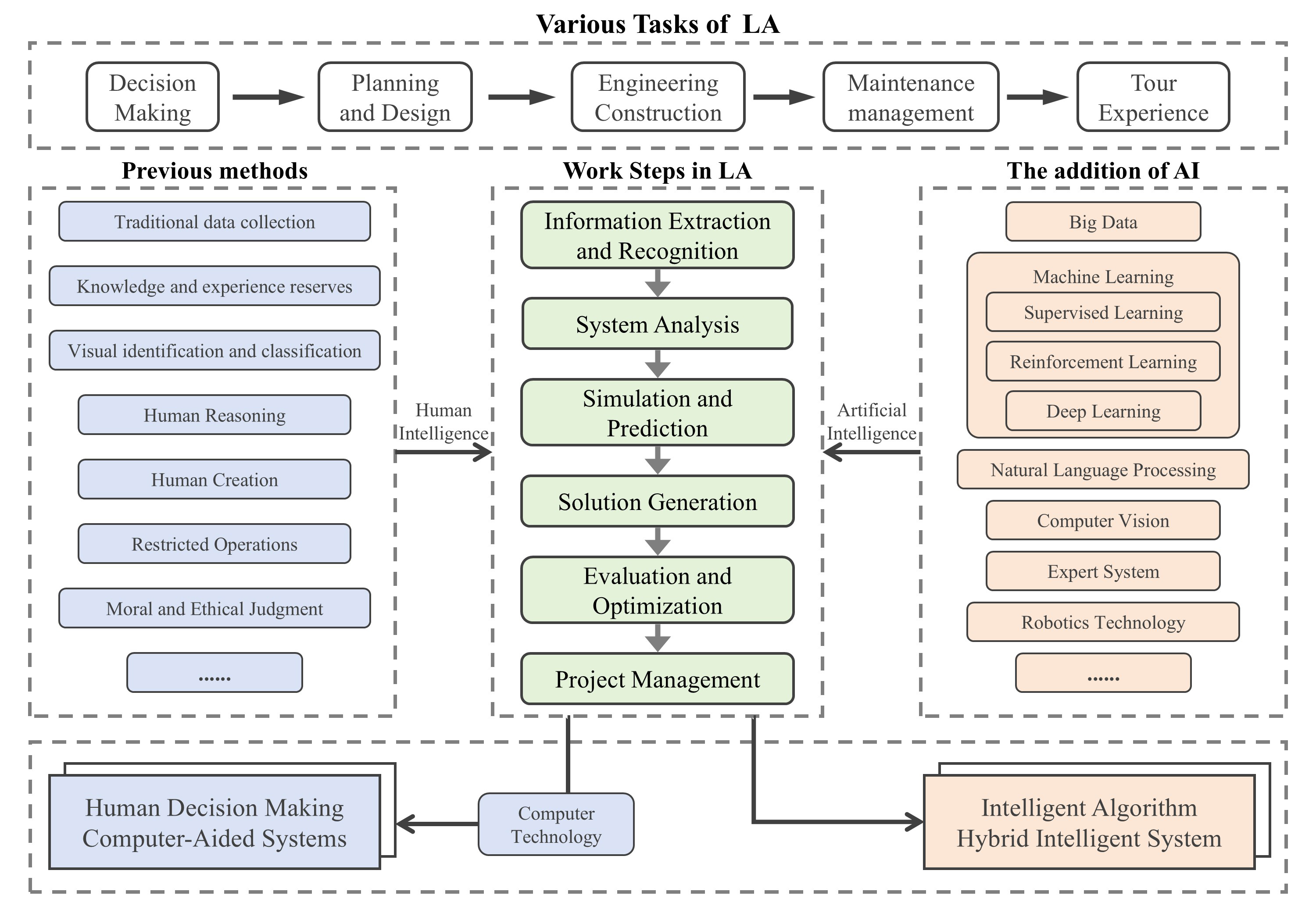}
    \caption{Various tasks and work steps in LA.}
    \label{fig:enter-label}
\end{figure}

\textbf{The connection between AI and LA:} In the research of LA, including planning, design, construction, maintenance, management, and other areas \cite{gazvoda2002characteristics}, with the development and application of AI technology, many issues can be effectively solved. Traditional LA workflow mainly relies on human intelligence \cite{korteling2021human} methods and computer technology assistance. The addition of AI and various new technologies has made this workflow very different. the common approach \cite{xu2021artificial} is to break down various tasks of LA into specific task objectives according to different work steps, and then design corresponding algorithms to solve the respective problems, as shown in Figure \ref{fig:enter-label}. Therefore, AI has a close relationship with the field of LA, and the following are several important application directions of AI in LA.

\textbf{(1) Landscape design and planning}: AI can assist landscape designers and planners in creative generation, scheme optimization, and decision-making support. Through machine learning and generative models, AI can analyze and learn from a large amount of landscape data and cases, providing design inspiration and automatically generating design schemes. Its workflow can be divided into stages such as site information extraction, landscape analysis and evaluation, project concept design, design scheme refinement, preliminary design, and construction drawing output. Especially in the initial design stage, a large amount of relevant information needs to be collected and organized, and a lot of technical issues need to be dealt with, requiring designers to have sufficient knowledge reserves. However, the emergence of AI technology can make the collection, storage, and analysis of landscape information more efficient and accurate \cite{oh2001landscape}, effectively solving similar problems; for example, land use identification and classification based on remote sensing and machine learning technology can scientifically, accurately, and efficiently analyze a large amount of data based on different classification targets and objects \cite{zhang2019joint,tong2020land}, essentially replacing the traditional visual identification and classification method, allowing designers to be free from the constraints of objective factors and greatly improving work efficiency, realizing the effective utilization of data. AI can also simulate and evaluate the impact of different landscape design schemes on the ecosystem. By establishing models and algorithms, AI can analyze and predict the impact of landscape design on aspects such as water resources, climate change, soil quality, and biodiversity \cite{christin2019applications}. These all help to optimize design schemes and ensure their ecological friendliness and sustainability.

\textbf{(2) Landscape architecture construction}: The active cooperation and coordinated management of various main bodies such as the management, construction, design, and construction parties play a decisive role in the construction of the project \cite{liu2017understanding}. As an important part of the infrastructure, due to the lack of communication, coordination, and information sharing among the various participating parties in the rapid development of the construction process, some problems such as arbitrariness, lack of professionalism, and poor engineering effects have occurred in the construction process. The AI technology has optimized and improved the entire process of landscape construction. For example, the LIM model can be used to extract material lists, carry out virtual construction, and coordinate the dynamic management of production factors such as personnel, materials, and machinery, generating dynamic reports on progress management, safety management, and cost management \cite{zhao2022intelligent}. Generative AI, such as AI-generated content \cite{fui2023generative,wu2023ai} can also automatically generate optimized construction plans by learning and analyzing a large amount of construction data and site conditions, helping project management personnel better plan and arrange the construction schedule \cite{kumar2015bim,kim2016multiobjective}, improving construction efficiency and reducing costs.

\textbf{(3) Landscape maintenance and management}: Decision-making, planning, design, construction, maintenance, and management in landscape construction are the full life cycle of landscape projects, and green maintenance and construction processes are equally important. Corresponding maintenance and management work must be done to ensure the survival rate of plants and enhance the social value and benefits of landscape projects. In the early stages of landscape planning and design, the lack of a holistic project perspective has led to unreasonable plant selection or collocation, single plant species, unclear layering, and a failure to consider the seasonal growth of plants, which has inadvertently increased the difficulty of green maintenance management and the cost of urban landscape construction \cite{zuo2014green}. However, AI applications can provide maintenance plan recommendations for plants, such as intelligent irrigation systems that can learn and analyze soil moisture, plant water, and nutrient requirements, and automatically generate intelligent irrigation and fertilization plans. Remote control of pumps and valves to achieve large-scale automatic irrigation \cite{bwambale2022smart} can improve the efficiency of green irrigation while reducing manpower input and water resource waste, effectively reducing the cost of later maintenance.

AI can improve the visitor experience and manage public spaces. Through an intelligent tour guide system, AI can recommend personalized tour routes and activities based on the interests and preferences of visitors \cite{buhalis2015smart}. It can provide real-time information about attractions, facilities, and activities to help visitors better understand and experience the landscape. AI can be used to monitor and manage the safety and environmental conditions of the landscape. Through image recognition and sensor technology, AI can monitor the flow of people, traffic conditions, garbage disposal, and safety risks in scenic areas, providing early warning and emergency response. It can help improve the safety and management efficiency of public spaces. These application directions demonstrate the potential and value of AI in the field of LA. To summarize, the various AI technologies in landscape architecture can provide more efficient, accurate, and sustainable design, planning, and management solutions, contributing to the creation of beautiful and livable urban environments.

\subsection{Application of AI in LA Design}

The diverse applications and interdisciplinary nature of AI technology, with its powerful data processing and rule-discovery capabilities, have effectively integrated into traditional landscape analysis, solving many planning and design challenges in LA. Therefore, this subsection introduces the AI technologies commonly used in LA research, their technical characteristics, and the LA problems that each type of technology can solve. It categorizes the existing AI application technologies into four key aspects: analysis and evaluation, planning and design, construction, and maintenance and management. Through specific simulation, data analysis, and causal analysis, it demonstrates the various possibilities of AI in solving LA problems. In addition, this article also briefly discusses other potential application technologies, including visualization simulation and landscape heritage protection. Details are shown in Table \ref{tab:applications}.

\begin{table*}[ht]
\small
\centering
\caption{Main technical applications of AI in LA }\label{tab:applications}
\scalebox{0.65}{\begin{tabular}{c|c|c|c|c|c|c|c|c|c|c|c}
\hline
\multicolumn{2}{c|}{\diagbox{AI technology}{Usage classification}} & \multicolumn{1}{l|}{\makecell{Land use\\and land\\Cover change}} & \multicolumn{1}{l|}{\makecell{Image\\recognition\\information\\extraction}} & \multicolumn{1}{l|}{\makecell{Text\\recognition\\and\\processing}} & \multicolumn{1}{l|}{\makecell{Ecosystem\\simulation}} & \multicolumn{1}{l|}{\makecell{Environmental\\forecast}} & \multicolumn{1}{l|}{\makecell{Landscape\\system\\analysis}} & \multicolumn{1}{l|}{\makecell{Design\\generation}} & \multicolumn{1}{l|}{\makecell{Landscape\\ evaluation}} & \multicolumn{1}{l|}{\makecell{Solution\\optimization}} & \multicolumn{1}{l}{\makecell{Construction\\and\\management}} \\ 
\hline
  \multicolumn{1}{c|}{\multirow{10}{*}{\rotatebox{90}{Machine learning}}} & {Naive bayes, NB} & \checkmark & & & & & & & & & \\
\cline{2-12} 
   & {Decision tree, DT} & \checkmark & & & & & & & & & \\
\cline{2-12} 
  & {Random forest, RF} & \checkmark & & & & & & & & & \\ 
\cline{2-12} 
  & {Cellular automata, CA} & & & & \checkmark & \checkmark & & & & & \\ 
\cline{2-12} 
  & {\makecell{Support vector\\machine, SVM}} & \checkmark & & & & & & & & & \\ 
\cline{2-12} 
  & {\makecell{Artificial neural\\network, ANN}} & & \checkmark & & & & & & & & \\ 
\cline{2-12} 
  & {\makecell{Recurrent neural\\network, RNN}} & & \checkmark & & & & & & & &\\ 
\hline
\multicolumn{1}{c|}{\multirow{12}{*}{\rotatebox{90}{\makecell{Natural language\\processing}}}} & \makecell{Bidirectional encoder\\representations\\from transformers, BERT} & & & \checkmark & & & & & & & \\ 
\cline{2-12} 
& \makecell{Generative pre-trained\\transformer, GPT} & & & \checkmark & & & & & & & \\ 
\cline{2-12} 
 & \makecell{Conditional random\\Field, CRF} & & & \checkmark & & & & & & & \\ 
\cline{2-12} 
 & \makecell{Long Short-Term\\memory, LSTM} & & & \checkmark & & & & & & & \\ 
\cline{2-12} 
 & \makecell{Agent-Based\\model, ABM} & & & & \checkmark & & & & \checkmark & & \\ 
\cline{2-12} 
 & \makecell{Multi-Agent\\system, MAS} & & & & \checkmark & \checkmark & & & & & \\ 
\hline
\multicolumn{1}{c|}{\multirow{5}{*}{\rotatebox{90}{\makecell{Computer\\vision}}}} & \makecell{Convolutional neural\\network, CNN} & & \checkmark & & & & & & & & \\ 
\cline{2-12} 
 & \makecell{Generative adversarial\\network, GAN} & & \checkmark & & & & & \checkmark & & & \\ 
\cline{2-12} 
 & CycleGAN & & & & & & \checkmark & \checkmark & & &\\
\cline{2-12} 
 & Extended reality & & & & \checkmark & & & & & & \\
\hline
\multicolumn{1}{c|}{\multirow{5}{*}{\rotatebox{90}{Optimization}}} & Swarm Intelligence & & & & \checkmark & & & & & \checkmark & \\ 
\cline{2-12} 
 & Genetic algorithms,GA & & & & & & & & & \checkmark & \\
\cline{2-12} 
 & Simulated annealing & & & & & & & & & \checkmark & \\
 \cline{2-12} 
 & Divide and conquer & & & & & & & & & \checkmark & \\
  \cline{2-12} 
 & Dynamic programming & & & & & & & & \checkmark & \checkmark & \\
\hline
\multicolumn{1}{c|}{\multirow{7}{*}{\rotatebox{90}{\makecell{Hybrid\\intelligent system}}}} & \makecell{Landscape information\\modeling, LIM} & & & & \checkmark & & \checkmark & \checkmark & & \checkmark & \checkmark \\ 
\cline{2-12} 
 & \makecell{Vegetation information\\modeling, VIM} & & & & & & & \checkmark & & \checkmark & \\ 
 \cline{2-12}
 & \makecell{Building information\\modeling, BIM} & & & & & & & \checkmark & & \checkmark & \checkmark \\ 
 \cline{2-12}
 & i-Tree Eco & & & & & & & & \checkmark & & \\ 
 \cline{2-12} 
 & DeepCity & & & & & & & \checkmark & \checkmark & & \\ 
\hline
\end{tabular}}
\end{table*}

\subsubsection{Ecosystem Simulation and Evaluation}

In various LA projects, the first step is the extraction, classification, analysis, and evaluation of site information, which is crucial for planning and design work. To solve practical problems more scientifically and efficiently, artificial intelligence technologies have been widely applied, with related research focusing on landscape pattern analysis and site ecological optimization, covering a variety of algorithms and models.

First, for data classification problems, machine learning (ML) can be used to analyze large amounts of ecological data \cite{stupariu2022machine}, and discover patterns and associations between data, including algorithms such as naive bayes (NB) \cite{rish2001empirical}, support vector machine (SVM) \cite{chang2011libsvm}, decision tree \cite{quinlan1986induction}, classification and regression tree (CART) \cite{breiman2017classification}, and random forest (RF) \cite{breiman2001random}. These algorithms have been widely used in landscape land use classification research. Among them, RF is often used in landscape ecology and land cover classification research \cite{cutler2007random,rodriguez2012assessment}, helping to effectively classify and analyze site information. Artificial neural networks (ANNs) \cite{jain1996artificial} have also been widely applied in landscape classification, simulation, prediction, and evaluation \cite{roman2020application}.

Secondly, for image information recognition and extraction problems, deep learning (DL) \cite{lecun2015deep}, a branch of machine learning, can process large-scale complex data through multi-layered neural network simulation of the human brain's working principles, and learn to represent features, such as convolutional neural networks (CNNs) \cite{gu2018recent}, recurrent neural networks (RNNs) \cite{schuster1997bidirectional}, and generative adversarial networks (GANs) \cite{goodfellow2014generative}. Due to their powerful image recognition capabilities, they are suitable for rapid identification and information extraction and can play a strong role in processing complex ecological data. In remote sensing imagery and street view image processing, deep learning algorithms can efficiently identify land cover types and landscape features \cite{li2020automated,xia2021development}. Deep learning models can be extended to larger regions, allowing for the analysis of multiple cities and regions in a short period \cite{zhang2018measuring}. CNNs also play an important role in landscape picture information recognition and feature extraction \cite{schiefer2020mapping,buscombe2018landscape}, providing important support for landscape pattern analysis.

For rich text data processing, many algorithms in the field of natural language processing (NLP), such as TF-IDF \cite{robertson2004understanding}, word2vec \cite{yao2017sensing}, bidirectional encoder representations from transformers (BERT) \cite{devlin2018bert}, conditional random fields (CRF) \cite{lafferty2001conditional}, and long short-term memory (LSTM) \cite{hochreiter1997long}, have been applied to text recognition and sentiment analysis, providing strong support for landscape perception pattern research and opinion analysis \cite{huai2022environmental}. These algorithms help classify the themes of text data on the internet and extract sentiment \cite{dang2020sentiment}, providing a more comprehensive perspective for comprehensive evaluation.

In addition, as tools for landscape system analysis and simulation, various tools have been applied at different levels of LA, architecture, and urban planning. For example, agent-based models \cite{abar2017agent} simulate the behavior of individuals in an ecosystem and study their interactions and impacts on the environment. They have been applied to design and simulate pedestrian flow and evaluate urban environments. Multi-agent systems \cite{dorri2018multi} can provide more quantitative data support through predictive simulation, offering more possibilities for landscape space development. Swarm intelligence algorithms can simulate group behavior activity trajectories, combined with space syntax analysis, to gradually explore and optimize solutions. They can improve the rationality of design schemes \cite{esposito2020agent} and have been widely used in LA planning and design.

Finally, for specific applications, there are a series of advanced technologies and systems that play an important role in evaluating and analyzing landscape characteristics, such as i-Tree Eco \cite{raum2019achieving}, a model for evaluating urban forest ecosystem services, which aims to use standardized field data from randomly located plots across the study area, combined with local hourly air pollution and meteorological data, to calculate ecosystem services based on tree census data in the region \cite{wu2019using}. i-Tree Eco is currently a complete method for quantifying the ecosystem services of urban trees in landscape ecology research. eCognition \cite{gupta2014object,yang2020multi} uses intelligent image analysis technology to improve the automatic recognition accuracy of spatial image data, and can also quantify vegetation landscape features. DeepCity \cite{dubey2016deep} can evaluate and quantify different urban forms, assisting designers in typological research on urban forms.

\subsubsection{Visualization Simulation and Prediction}

In landscape architecture, visualization simulation not only can evaluate the impact of landscape design on the surrounding environment, but also can help designers and planners better understand the characteristics and constraints of the site \cite{bergen1998data}, and optimize and adjust the spatial layout during the design process, ultimately allowing them to present their concepts and design ideas to clients, team members or decision-makers more intuitively and vividly. In recent years, AI technology has realized visualization simulation and prediction in LA through the application of various algorithms.

In data analysis and modeling \cite{gan2021fast,gan2021utility}, regression algorithms can automatically associate data and perform function fitting to uncover the inherent connections behind the data \cite{fu2023enhancing}. Algorithms such as principal component analysis (PCA) \cite{abdi2010principal} and logistic regression \cite{hosmer2013applied} can analyze and model various data to explore the correlation between data and reveal the underlying connections. This data may involve multiple aspects such as landscape patterns, geographic information, and environmental data. For example, by analyzing historical landscape data, it is possible to predict the development of future landscape patterns \cite{turner1989landscape}. On this basis, algorithms such as cellular automata (CA) \cite{wolfram1984cellular} and multi-agent systems (MAS) \cite{ferber1999multi} have powerful simulation capabilities and play an important role in the simulation and prediction of future landscape patterns, among which CA can simulate dynamic systems such as urban growth and landscape change \cite{liu2014simulating}, and can present the inherent mechanisms and rules of landscape evolution. Humans can predict the trend of future landscape development through the simulation results of CA, providing reliable support for landscape planning and design. In 2007, Herr and Kvan \cite{herr2007adapting} proposed a generative architectural design process using CA.

For prediction guidance and optimization, the decision tree can flexibly adjust according to constraints and generate multiple hypotheses, providing predictive guidance for landscape design \cite{raman2006computer}. This provides designers with a reference for the prediction results, guiding the optimization and adjustment of design schemes. In addition, swarm intelligence algorithms \cite{kennedy2006swarm} simulate the trajectory of group behavior activities and combine them with spatial syntax analysis to further interpret the simulation results. These analysis results can provide a basis for scheme optimization \cite{johnson2002emergence,hillier2007space}, helping designers more effectively explore the best schemes and improve the rationality of landscape design.

For visualization display and interactive experience, through virtual reality technology, the simulation and prediction results can be visualized, and users can enter the virtual environment to watch and manipulate the virtual world generated by the computer, realizing an immersive interactive experience \cite{portman2015go}. This virtualized interactive approach can provide a new medium for landscape presentation, helping designers and decision-makers more intuitively understand the simulation results and provide feedback and adjustments.

\subsubsection{Landscape Planning and Design}

Due to the powerful image recognition and generation capabilities of deep learning technology, in the field of landscape planning and design, it can generate similar data of the same type by learning a large amount of case data, such as 2D images and 3D data, and is often used in the generation of landscape design schemes, providing designers with more possibilities and inspiration.

GAN is an important algorithm in deep learning, which can generate realistic images through adversarial training \cite{isola2017image}. Among them, the Pix2Pix model of GAN can automatically generate site layout designs and generate more diverse plan results \cite{huang2018architectural}. These schemes not only follow the principles of garden design but also consider the aesthetic visual effects based on reasonable spatial layout. The application of parametric generation technology can effectively improve the scientificity of planning and design, making model modification more convenient and saving a lot of time \cite{luo2021online}. Cycle generative adversarial network (CycleGAN) \cite{ye2022masterplangan} can realize the extraction of different land use types on the plan, as well as the rendering generation from the plan color block diagram to the color texture diagram, thereby improving the analysis and mapping efficiency of designers.

Digital twin technology \cite{liu2021review,ye2023developing,tan2024digital} can digitally reproduce the actual environment, it can simulate the real environment and its changes in real-time, providing a foundation for the realization of intelligent planning and design. Digital twin technology can help designers better understand and simulate the landscape environment, thereby optimizing the design scheme, and interacting with the real system in real-time, providing more intuitive design reference and decision support. DeepCity \footnote{DEEPCITY open source software download website: \url{https://github.com/kekehurry/DeepCity}} is a digital design tool that can automatically learn the morphological patterns of the urban fabric types specified by the designer and apply them to new urban environments. In addition to the design generation part, DeepCity also includes a design evaluation function, which can quickly evaluate the physical performance of the design scheme in the early design stage, thereby assisting the designer in modifying and deepening the scheme. This tool uses digital technology to provide more intelligent and efficient design support.

Landscape information modeling (LIM) \cite{ahmad2012need,oh2001landscape} has cross-scale data fusion functions in planning and design, and can provide professional data analysis for national spatial planning, ecological space planning, rural landscape planning, etc. LIM uses digital technology to assist in planning and design, design deepening, and scheme simulation applications, enhancing the scientificity and efficiency of design.

\subsubsection{Landscape Architecture Construction} 

The building information modeling (BIM) platform \cite{azhar2011building,volk2014building} is an integrated platform for the entire life cycle of construction. Through the interconnection and information sharing of construction data, it provides a complete 3D model and construction information library, offering accurate reference and information for designers and constructors to improve design quality and construction efficiency.

EPC project management and supervision system \cite{dzhusupova2022challenges} is a management model that deeply integrates survey, design, procurement, and construction, solving the conflicts between design and construction management. It provides comprehensive engineering services throughout the entire construction process, which can also be assisted by AI technology for project management and supervision.

Different virtual reality (VR) and augmented reality (AR) technologies in LA \cite{portman2015go,wang2013conceptual} can construct the Metaverse (both virtual and reality) scenarios \cite{sun2022metaverse,chen2024open}. They use immersive techniques to optimize design schemes, guide construction through visualization, and use digital twin technology \cite{ryzhakova2022construction} to detect and provide feedback on project information. This not only improves design efficiency but also provides visual guidance for construction.

Generative AI \cite{barcaui2023better,wu2023ai} can optimize construction plans, monitor and control construction processes, and achieve construction quality assessment. For example, AI can generate smart irrigation and fertilization plans based on plant needs and environmental conditions to improve water resource utilization and plant growth quality.

LIM \cite{zhao2022intelligent,wik2018bim} can be used to extract material lists, conduct virtual construction, and dynamically manage production factors such as personnel, materials, and machinery, generating dynamic reports on progress, safety, and cost management. The integration of LIM model data and automatic control of construction machinery has realized the automatic construction of earthworks and the informatization management of seedlings, providing support for smart construction sites and digital twin construction sites, thereby improving construction efficiency and quality.

\subsubsection{Landscape Maintenance and Management}

The intelligent maintenance plan \cite{wolfert2017big} plays an important role. Through generative AI, it can automatically generate suggestions for greening areas, plant species, and layouts, helping to rationally plan the city's greening layout and improve the ecological environment and residents' quality of life. Plant disease and pest identification \cite{ferentinos2018deep} is also an important part of intelligent maintenance, where AI can identify plant species, diseases, and pests, and analyze vegetation coverage and growth status using computer vision technology. In addition, through sensor networks, it can monitor environmental parameters such as air quality and noise levels, learn and identify various environmental sensor data, and perform real-time monitoring and analysis of the park environment to help quickly respond to and address environmental issues, protecting the ecological environment and residents' health.

The intelligent irrigation system \cite{goap2018iot} is a key technology in landscape maintenance. Combining modern automatic control technology, data analysis and processing technology, and communication technology, the system can remotely control pumps and valves to achieve unmanned automatic irrigation, which not only improves the efficiency of greening irrigation and reduces labor input, but also helps conserve water resources. Furthermore, the integration of intelligent control technology and information technology makes it easy and quick to remotely view, operate, control, and maintain the irrigation system, further improving management efficiency and the accuracy of irrigation control.

The intelligent lighting system \cite{de2016intelligent} is another important technology application. Through AI technology, the system can intelligently adjust the lighting based on on-site lighting intensity, temperature, and humidity, creating a coordinated lighting solution for the actual environment. The system is also equipped with sensors to enhance the interactivity and entertainment of the lighting, providing users with a high-quality visual experience while creating an enjoyable landscape environment for relaxation and exercise.

Finally, smart energy technology \cite{lund2017smart} also plays an important role in park management. This technology, through the comprehensive application of AI, IoT, and big data, integrates the energy system and information system, realizing the intelligent management and control of energy, efficient utilization of energy, optimization of energy structure, and reduction of energy consumption and pollution, providing sustainable energy support for landscape maintenance and management.

\subsubsection{Landscape Heritage Protection}

AI technology can be applied to the digitization of information collection, organization, and processing, providing comprehensive support and assistance for landscape heritage protection, thereby realizing the scientific and effective protection and management of landscape heritage. For example, through deep learning algorithms, such as convolutional neural networks (CNN) \cite{abed2020architectural}, it is possible to automatically identify and analyze landscape elements such as historical buildings, cultural relics, and historic sites, and store their information in a digital archive. The digitization of cultural landscape heritage \cite{wang2017application} is to collect various information about the cultural landscape through digitization and unify its storage, analysis, and visualization, thereby establishing a scientific heritage archive. Research \cite{wang2022artificial} has pointed out that the digitization of traditional village cultural elements and their display and dissemination in digital form can help promote public understanding and participation. This digital protection has brought unprecedented opportunities for the inheritance and protection of traditional village culture.

Both data mining \cite{huang2024taspm} and pattern recognition technologies can be used to analyze digital information in-depth and uncover the patterns and characteristics hidden in the data. This can help understand the evolution process of landscape heritage and evaluate potential threat factors. For example, using natural language processing (NLP) \cite{sporleder2010natural,dou2018knowledge} to digitally process and analyze literature, including the organization and archiving of historical documents, archaeological reports, and expert opinions, to provide support for understanding the historical and cultural background of landscape heritage. In addition, various AI applications in virtual reality (VR) and augmented reality (AR) \cite{zhang2019cityscape,zhong2021application} can also help improve the public's awareness and understanding of landscape heritage. By recreating, simulating, and displaying landscape heritage, people can immerse themselves in the experience of landscape heritage, promoting conservation awareness and participation, as shown in Table \ref{table:Mian technologies}.

Big data technology \cite{wu2013data,gan2017data} can also unlock potential value in the field of LA, helping designers and planners make more informed decisions. For example, big data provides information on demographics and user behavior \cite{bentley2014mapping,sun2023internet}, helping designers understand the needs and preferences of target users. By analyzing demographic data, social media data, and mobile device data, insights can be gained about population movement, activity patterns, and usage habits. This information can guide the layout of facilities, activity area planning, and public space design in landscape design. Big data analysis can also provide climate and environmental data \cite{guo2015earth}, including temperature, precipitation, and wind direction, which are crucial for plant selection, irrigation management, and landscape sustainability assessment. By analyzing historical climate data and real-time weather monitoring, designers can better understand the climate conditions and provide scientific support for plant configuration and water resource management. In general, big data in LA can provide more comprehensive and accurate information to help designers make more informed decisions and enhance the sustainability, aesthetics, and user experience of landscapes. However, data privacy and security are also issues that need to be considered and addressed, to ensure the legal collection, processing, and protection of data.

\subsection{The Practice of AI in LA}

This section introduces LA projects that have applied AI technology, categorized into four scenarios: landscape planning and design, maintenance and management, landscape heritage preservation, and tourism landscape guidance.

\textbf{(1) AI application in landscape planning and design}. AI can be used to assist landscape designers in generating creative ideas and optimizing design schemes. By analyzing a large amount of landscape data, images, and geographic information, AI can provide design inspiration, automatically generate landscape layouts, and select plants based on environmental requirements. These tools can improve design efficiency and quality while reducing human errors. The application of parametric generation technology has enhanced the scientific support of landscape design. In terms of platform selection, geographic information system (GIS) \cite{wong2005statistical} can be used for preliminary data analysis of large-scale terrain, landforms, slopes, and aspects. Grasshopper \footnote{Grasshopper website: \url{https://www.grasshopper3d.com/}} running on the Rhino platform has advantages in the design of complex curved surfaces, which can be used for terrain, landscape architecture modeling, and the design of complex surface textures of irregular structures. \textbf{Specific case (a)}: The urban renewal planning practice project in Dengfeng city \cite{Zheng2020Research_CN}. Based on fully convolutional networks (FCN) \cite{chen2019aticnet} and a deep learning algorithm-based urban scene element dataset, this project accurately identified various elements in the urban landscape environment of Dengfeng City. By training a large dataset of landscape element recognition, the recognition accuracy was improved, and the urban landscape problems were precisely diagnosed. The results of intelligent recognition provided scientific support for the design practice, analyzing the current landscape problems in Dengfeng, and refining the urban built environment to improve the living environment quality. \textbf{Specific case (b)}: The construction of digital twin platform for rural ecological landscape \cite{tan2024digital}. The project used a digital twin rural ecological landscape digitalization platform to perform integration, analysis, and simulation of the ecological processes and spatial forms in the landscape space, as well as the reproduction of the landscape ecology. Digital twin technology not only constructed a digital landscape environment but also loaded and integrated multi-source heterogeneous data such as text information, geographic information, spatial modeling, images, audio, and sensor data, providing richer and more comprehensive information to build a future smart map.

\begin{strip}
  \centering
    \captionof{table}{Main technologies of AI in LA design} 
    \label{table:Mian technologies} 
  \tablefirsthead{ %首页表头
  \hline \multicolumn{1}{c|}{\textbf{Technology}} & \multicolumn{1}{c}{\textbf{Content}} \\}
  \tablehead{ %后续页表头
  \hline \multicolumn{2}{l}{\small\sl Continued from previous page}\\
  \hline \multicolumn{1}{c|}{\textbf{Technology}} & \multicolumn{1}{c}{\textbf{Content}} \\}
  \tabletail{ %除最后一页外的表尾
  \hline \multicolumn{2}{r}{\small\sl Continued on next page}\\ \hline}
  \tablelasttail{\hline} %最后一页的表尾
\small
\begin{supertabular}{m{2.8cm}<{\raggedright}|m{13.7cm}<{\raggedright}}
    \hline
    
    Artificial neural network (ANN) \cite{jain1996artificial} & It is widely used in the fields of landscape classification, simulation prediction, and evaluation \cite{roman2020application}. ANN is inspired by biological nerves and brain structures. A large number of original technologies can be used to solve landscape practice problems, which are manifested in the use of activation functions and multi-hidden layer structures, greatly enhancing learning ability, expression ability, and associative memory ability. \\  \shrinkheight{-20ex}  \hline  
    
    Agent-based model \cite{esposito2020agent,abar2017agent} & By simulating the individual behaviors in the ecosystem, studying their interactions and their impact on the environment, it is applied to the design and simulation of human flow, evaluation of urban environment, etc. \\   \hline

    Cellular automata (CA) \cite{wolfram1984cellular,herr2007adapting,liu2014simulating,zhai2020simulating} & It is widely used in dynamic system simulation and modeling of urban growth processes, landscape replacement, spatial ecology, and land dynamics, etc., showing the internal mechanisms and laws of landscape evolution, predicting the future development of landscape, and providing guarantees for management and planning. \\  \hline

    Convolutional neural networks (CNN) \cite{gu2018recent,schiefer2020mapping} & It can automatically learn from massive images and extract abstract features, so it is widely used in landscape image information recognition, feature extraction, and classification. \\      \hline

    CycleGAN \cite{ye2022masterplangan} & This training model has the potential to be applied to landscape land use type analysis and plane rendering, improving analysis and mapping efficiency, and is widely used in processing image classification, image style transfer (combining the content of one image with the style of another image), and image restoration. Compared with CNN, GAN's recognition and generation training can be completed in one model, and the model can complete classification and generate new images after training. \\  \hline 

    Decision tree \cite{quinlan1986induction,raman2006computer} &  It can produce a decision tree with strong generalization ability and can handle unseen examples. Its basic process follows a simple and intuitive "divide and conquer" strategy. The algorithm can adjust according to the constraints and generate multiple hypotheses, which can provide predictive guidance for the behavioral outcomes of landscape design. \\       \hline

    Deep learning and panoramic image technology \cite{lecun2015deep,li2020automated,christin2019applications} & It has a wide range of applications in plant landscape quantification. Commonly used plant landscape evaluation indicators can be processed through deep learning technology, which has strong adaptability. As landscape evaluation continues to develop digitalization, it has shown outstanding results in cases with too large a sample size. \\     \hline

    DeepCity & It can recognize, generate, and evaluate designs. It can be used to evaluate and quantify different urban forms and conduct typological research on urban forms. It can automatically learn the morphological patterns of urban texture types and apply them to new urban environments to weave and repair specific types of urban forms. It can quickly evaluate the physical performance of design solutions and assist in modifying and deepening the solutions in the early stages of design. \\     \hline

    Digital construction of urban landscape system & From the perspective of regional design, a series of technology clusters are proposed to deal with the mountain-sea urban landscape system in combination with digital technology, including regional digital landscape patterns, regional digital habitat network, regional digital mountain-sea context, and regional digital mountain-sea style. Through the four-in-one urban design digital analysis method, the mountain-sea urban landscape system is optimized and designed at the regional level. \\    \hline

    i-Tree Eco \cite{raum2019achieving,wu2019using} & Using standardized field data from randomly located plots within the study area, combined with local hourly air pollution and meteorological data, to calculate ecosystem services based on tree census data within the area is currently a relatively complete method for assessing urban tree ecosystem services. \\   \hline

    Digital dynamic landscape & It is different from the static form of traditional landscape. Virtual digital landscapes can bring immersive, interactive, and positive dynamic experiences, and render the landscape space atmosphere through light effects, sounds, and virtual scenes, which increases the perception dimension of the experience and makes the individual feel sensory pleasure when participating in the activity. Digital dynamic landscape breaks through the traditional landscape form. The sensory shock people feel comes from the immersive, dynamic experience, lighting, and material rendering of the virtual landscape. \\   \hline

    Digital landscape & The system simulates and reflects the ecology and morphology of the landscape environment and their interrelationships. Digital landscape theory, methods, and technologies can realize the full process digitization of complex data acquisition, cognitive analysis, planning and design, scheme selection, construction, operation, and maintenance management, thereby reducing errors caused by human intervention or subjective judgment. \\   \hline   

    Digital recording of cultural landscape heritage \cite{wang2022artificial,cai2021joint,liu2020pattern} & It has five major characteristics: comprehensive, accurate, dynamic, integrated, and open. To capture and reflect the characteristics and value of heritage, it uses digital means to collect various types of information on cultural landscapes and uniformly stores, analyzes, and visualizes them, thereby establishing a scientific heritage archive and providing sufficient and timely information support for protection and management. \\     \hline
    
    Digital twin (DT) \cite{liu2021review,ye2023developing,ryzhakova2022construction} & Using digital twin technology to map the interaction between landscape ecology and morphology and its changing mechanism, explore the intrinsic relationship between landscape ecology and morphology, collect and integrate multi-source heterogeneous data, analyze and evaluate the dynamics of landscape environment, and construct multi-information models. \\   \hline

    eCognition \cite{gupta2014object,yang2020multi} & Intelligent image analysis adopts a fuzzy classification algorithm supported by a decision-making expert system and proposes an object-oriented analysis classification technology to effectively extract information on typical vegetation landscape elements, improve the automatic recognition accuracy of spatial image data, and provide a new type of quantitative analysis technology for the study of vegetation landscape characteristics in scenic environments. \\      \hline    
        
    Extended reality (XR) \cite{portman2015go,rauschnabel2022xr} & Combining the real and the virtual, creating a landscape environment with human-machine interaction and virtual-real coexistence, integrating the visual interaction technologies of AR, VR, and MR (Mixed Reality), constructing a deep landscape scene, mapping the human living environment system and its process, and forming an "immersive feeling" with seamless conversion between virtual and real. \\   \hline

    Green and intelligent landscape construction full process management \cite{dzhusupova2022challenges} & EPC project management and project supervision is a management model that deeply connects survey, design, procurement, and construction. It resolves the contradictions in design and construction management and provides comprehensive engineering services for the entire process of design and construction. This can also be done with the help of artificial intelligence technology for project management and supervision. \\       \hline

    %General packet radio service, GPRS & Intelligent lighting system, latitude and longitude control instruments and time control instruments, wireless landscape lighting remote monitoring equipment. Adjust the lighting in time according to the on-site light intensity, temperature, humidity, and other environmental factors to form a lighting combination lighting plan that is coordinated with the actual environment and creates a good visual experience. The supporting intelligent sensing system, with built-in sensors and other devices, can sense environmental characteristics in a timely manner and automatically adjust parameters such as light color and intensity, enhancing the interactivity and fun of lighting, creating a very interesting landscape environment, so that people who exercise or relax in the landscape can release their emotions and soothe their mood. \\   \hline    

    HUL concept technology \cite{dahlhaus2014visualising} & In the HUL protection dynamic monitoring technology method, in the context of smart city construction, digital information technology is used to accurately and efficiently manage the changes in the core content of protection. Based on unit management, it further integrates the urban planning system and dynamically monitors the urban historical landscape. \\     \hline

    Hypertext preprocessor (PHP) & PHP is one of the most popular development languages at present. It has the advantages of low cost, high speed, good portability, and a rich built-in function library. Through this server, users can browse the landscape information of different historical stages and provide users with clear and interactive historical data retrieval tools. At present, the application is still being optimized and upgraded.\\   \hline

    Informal green space digital identification technology \cite{zhang2019joint,tong2020land,li2020automated,yang2020multi} & It can effectively reduce time and labor costs and expand the scope of recognition. Compared with manual visual interpretation, deep learning models can analyze and learn from large amounts of data, making recognition relatively objective, and the recognition results have higher precision, accuracy, and recognition efficiency. It can analyze multiple cities and regions in a short period of time. \\   \hline

    Landscape construction BIM platform \cite{liu2017understanding,kumar2015bim,volk2014building,azhar2011building} & It is an integrated platform for the entire life cycle of construction. It can provide a complete three-dimensional model and construction engineering information database for landscape construction through the interconnection, exchange, and sharing of construction information, and provide accurate reference and information for designers and builders to improve design quality and construction efficiency. \\       \hline

    Landscape information modeling (LIM) \cite{zhao2022intelligent,oh2001landscape,ahmad2012need,wik2018bim} & It uses BIM for LA, and its objects are mainly engineering projects, including the entire life cycle of projects such as planning, design, construction, and operation. The connection between LIM model data and automatic control construction machinery has realized the automatic construction of earthwork projects and the information management of seedlings, providing support for smart construction sites and digital twin construction sites, thereby improving construction efficiency and quality. \\     \hline

    Machine learning (ML) \cite{alpaydin2020introduction,stupariu2022machine,jordan2015machine} & Including data, algorithms, and application platforms, it is good at summarizing rules in various data and solving different specific problems in the three stages of information extraction, analysis, and evaluation, as well as planning and design in the landscape planning and design workflow. \\     \hline
        
    Multi-agent system (MAS) \cite{ferber1999multi,dorri2018multi} & Through predictive simulation system analysis, more possibilities are given to landscape space development, and more quantitative data is provided. The advantage is to use reasoning, simulation, and other methods to analyze the exposed or given LA knowledge and independently select appropriate solutions. \\    \hline 

    Parameterized generation \cite{oh2001landscape,luo2021online} & The application of this technology improves the scientific nature of planning and design, makes model modification more convenient, and saves a lot of time. In terms of platform selection, designers can use different parametric design software according to their task settings, such as using GIS for early design data analysis, including some large-scale terrain, landform, slope, and slope analysis; using GRASSHOPPER to model landscape garden terrain and regional landscape sketches, and also give special-shaped structures complex surface textures. \\    \hline    

    Pix2Pix model \cite{huang2018architectural} & As one of the models of adversarial generative networks, it can realize automatic layout design for blank sites, and can also generate more diverse plane results through diffusion models. The generated scheme not only conforms to the principles of garden design but also has reasonable spatial layout and aesthetic effects. \\     \hline

    Random forest (RF) \cite{breiman2001random,cutler2007random,rodriguez2012assessment}, & It uses small-scale training samples and limited computing resources to achieve accurate classification in a short time and improve the accuracy of problem prediction. It is a representative of ensemble learning algorithms. Its high-precision tree classifier features are often used in classification studies such as landscape ecology and land cover. The RF algorithm for remote sensing data extraction is widely used. \\       \hline
    
    Smart energy technology \cite{lund2017smart} & It is a comprehensive application technology based on the Internet of Things, AI, big data, etc., with the goals of efficiently utilizing energy, optimizing energy structure, reducing energy consumption and pollution, etc., organically combining energy systems with information systems to achieve intelligent management and control of energy. \\         \hline        

    Smart irrigation system \cite{bwambale2022smart,ullo2020advances,goap2018iot} & By combining irrigation technology with intelligent control and information technology, on-site and remote query, operation, control, and maintenance of the intelligent irrigation system are realized. Data is collected using front-end sensors, and intelligent decision-making is made using system modeling and data linkage, which reduces manpower input and water resource waste and improves management efficiency and accuracy. \\     \hline

    Smart LA & Combining human rational judgment with intelligent carriers, by establishing a large database, it can efficiently and accurately collect, store and analyze landscape information, and combine human wisdom with the beauty of nature to create a realistic landscape with aesthetics and science that transcends time and space. And use wisdom to reproduce, perceive, and experience the landscape, creating a realistic landscape with perceptual experience that transcends time and space. \\     \hline    
        
    Swarm intelligence behavior simulation \cite{kennedy2006swarm} & By digitally translating, simulating, and predicting the behaviors that will occur on the site, the humanized design of the park can be effectively assisted. The swarm intelligence algorithm simulates the trajectory of group behavior activities, feedbacks the design plan problems, combines space syntax analysis, interprets the simulation results from the perspective of the overall form of the plan, and gradually obtains the ideal optimization plan through plan adjustment and re-simulation to improve the rationality of the design. \\      \hline
        
    Virtual reality and augmented reality \cite{zhang2019cityscape,zhong2021application,wang2013conceptual} & Virtual reality technology, by constructing virtual reality scenes and using immersive technology to optimize design solutions, allows engineering projects to be guided in construction under visualization and uses digital twin technology to detect and feedback project information, providing more implementation basis for construction. This not only improves design efficiency but also provides visual guidance for construction. \\      \hline
        
    Virtual tour & Using computer integration, sensor measurement, simulation, microelectronics, and other related technologies to design a virtual environment, allowing the audience to be in a real-time three-dimensional virtual environment and be able to view and manipulate the virtual world generated by the computer. Hear the real sounds of your garden, smell the plants, and perceive and interact in the virtual environment. \\   
\end{supertabular}
\end{strip}

\textbf{(2) AI application in landscape maintenance and management}. The smart plant environment monitoring and irrigation system \cite{ullo2020advances} uses various sensors, such as soil monitors, climate monitors, light sensors, and temperature sensors, to detect soil moisture, pH, and porosity, detect harmful gases and carbon dioxide in the air, and monitor light intensity and temperature. Designers and managers can observe these monitoring results on computers and mobile apps. Dynamic plant growth models, sunlight change models, and plant water models can also be established to analyze the growth and soil moisture conditions of different plants. \textbf{Specific case (a)}: A case study of bird species in the Mount Lofty Ranges, South Australia.\cite{westphal2007optimizing} This project is an optimization case for landscape configuration: determining the optimal landscape restoration for 22 bird species in the Mount Lofty Ranges of South Australia, to maximize the number of species occurrences through vegetation restoration. This project provided one of the first applications of decision modeling tools for optimal habitat restoration in a real landscape, combining species-specific suitability functions. \textbf{Specific case (b)}: CityTree \footnote{\url{https://en.wikipedia.org/wiki/CityTrees}} - A smart greening solution. Green City Solutions, a German company, has developed a smart greening solution called "CityTree", which is a vertical green wall device that uses mosses and plants to purify the air and uses smart sensors to monitor air quality, temperature, and humidity. AI technology is used to optimize plant selection and configuration, and automatically adjust irrigation and lighting based on environmental conditions.

\textbf{(3) AI application in landscape heritage protection.} Technologies such as geographic information systems and machine learning have effectively improved the efficiency and accuracy in the identification of landscape features and elements, providing a data platform for the extraction, protection, and management of historical and cultural landscape heritage. \textbf{Specific case (a)}: Digitalization of the historic landscape in Ballarat \cite{buckley2015using}. As the first city in the world to systematically apply the Historic Urban Landscape (HUL) approach \cite{dahlhaus2014visualising} to guide the digitalization of heritage protection, Ballarat has established a digital heritage information service platform centered on the "Ballarat Historic Urban Landscape Network", which comprehensively monitors and shares information on the current urban environment. The main technologies include a web-based geographic information system platform, databases for urban heritage, natural environment, landscape characteristics, historical sightlines, and public facilities, as well as interactive applications designed with open-source scripting languages and collection of multimedia GIS-based HUL public data \cite{wu2024geospatial}. The innovation lies in strengthening the human-landscape interaction, where the information service system is designed according to the needs of different stakeholders in terms of content, format, and dissemination channels, providing customized information to different users, and promoting public participation in heritage protection decision-making. \textbf{Specific case (b)}: Digitalization of Mount Lushan cultural landscape heritage sites \cite{cai2021joint}. Mount Lushan is located in Lushan County, Jiujiang City, Jiangxi Province, China. It has a rich cultural and natural heritage and is famous for its beautiful and unique mountains, rivers, and lakes. It also contains landscape elements such as ancient buildings, villas, stone carvings, alpine plants, waterfalls, and streams. The team used an integrated process of oblique aerial photography, 3D laser scanning, and 360-degree panoramic technology. Oblique photography images are obtained through UAV, 3D models are built by creating point clouds, and the scanned and measured data are processed through registration, fusion, stratification, etc., with extremely high model accuracy. The Internet of Things (IoT) and analytical models \cite{sun2023internet} are integrated to build a virtual tourism system to provide users with a virtual experience of cultural landscape heritage tourism. \textbf{Specific case (c)}: Digital landscape of the Baojiazhuang ancient village in Anshun, Guizhou \cite{liu2020pattern}. This project is based on the automated identification of landscape elements using machine learning. Due to the complex landscape environment and structure in the region, including karst landforms, ancient water conservancy projects from the Ming Dynasty, as well as mountains, forests, farmlands, and residential buildings, the project first used digital photogrammetry modeling and machine learning techniques to automatically identify landscape elements and patterns, constructing a rural landscape image database as a data foundation for the intelligent management of rural land, landscape, and natural environment. The automated recognition based on machine learning achieved an accuracy of 79\% to 90\%, and further analyzed the spatial relationships between natural elements, improving the accuracy and efficiency of heritage investigation and analysis.

\textbf{(4) AI application in landscape interpretation for tourism}. AI can provide intelligent tour guide systems and interactive experiences to enhance visitors' experience in scenic areas. Using data mining and natural language processing technologies, personalized tour guidance and recommendations can be provided to visitors based on their interests and preferences. In addition, computer vision techniques can be used to realize scenic spot recognition and augmented reality experiences. London TreeTalk app \footnote{\url{https://www.treetalk.eco/}}, which is based on urban tree data, is an example that creates a personalized urban forest and green network map for residents, connecting them with the local natural environment and green infrastructure. These cases demonstrate the diverse applications of AI in landscape design, from greening solutions to visualization tools and intelligent systems, providing designers with more tools and resources to create more beautiful, sustainable, and intelligent landscape environments.

\section{Challenges and Opportunities in LA} \label{sec:challenges}
\subsection{Challenges of LA}

AI has many opportunities in the field of LA, such as creating and optimizing design solutions, broadening design thinking, and providing more creative possibilities for designers. Accurate data analysis and decision support - AI can provide decision support through data analysis and modeling. It can predict and evaluate the effects of different design schemes, optimize resource utilization and environmental impact, and provide scientific basis and recommendations. Ecological simulation and sustainability assessment provide ecosystem monitoring and early warning functions, helping to adjust management measures in a timely manner. Intelligent maintenance and management can improve management efficiency and resource utilization, and provide personalized plant maintenance solutions. Smart guidance and interactive experiences can recommend personalized tour routes and activities based on visitors' interests and locations, enhancing visitor engagement and satisfaction. These opportunities make AI have broad application prospects in LA. It can improve design efficiency, optimize resource utilization, support decision-making, and enhance user experience, helping to create more beautiful, sustainable, and intelligent garden environments.

AI in the field of LA also faces some problems and challenges, including but not limited to the following aspects:

\begin{itemize}
    \item \textbf{Data acquisition and quality} \cite{zhu2005cost}: AI requires a large amount of high-quality data for training and analysis. However, acquiring data related to LA may be somewhat challenging, especially in terms of geographic information, vegetation data, and ecosystem monitoring. In addition, the quality and accuracy of the data are crucial to the effectiveness of AI algorithms, so ensuring the accuracy and completeness of the data is a challenge.

    \item \textbf{Complexity and uncertainty} \cite{wu2020managing}: Landscape design and management involve multiple complex factors, including terrain, climate, vegetation, soil, etc. The interaction and uncertainty of these factors pose challenges for AI algorithms in making predictions and optimizations. Designers and managers need to carefully consider these factors and incorporate them into AI models and systems.

    \item \textbf{Human creativity and subjectivity} \cite{amabile2020creativity}: LA design is a creative and artistic work involving human aesthetics and subjectivity. Although AI can assist designers in generating creative ideas and optimizing design schemes, it cannot completely replace human creativity and subjective decision-making. Therefore, how to find a balance between AI technology and human creativity is a challenge.

    \item \textbf{Privacy and security} \cite{elliott2022ai}: The application of AI in LA may involve the collection and processing of personal privacy and sensitive data. Ensuring the privacy and security of this data is an important challenge, requiring appropriate data protection and security measures to safeguard user rights and data security.

    \item \textbf{Acceptance and educational promotion} \cite{zhang2021ai}: The application of AI technology LA is still relatively new, and there may be unfamiliarity and resistance to new technologies among people. Both relevant technology education and promotion work are crucial to improving people's acceptance and understanding of AI applications.
\end{itemize}

\subsection{Challenges of LA in China}

These challenges need to be solved through continuous research and practice. With the progress of technology and the accumulation of experience, we can expect AI to play a greater role in LA and overcome the current challenges. Compared to developed countries, there are some gaps in landscape planning in China:

\begin{itemize}
    \item \textbf{Experience and history}: Some developed countries have a long history and rich experience in landscape planning. For example, the urban parks and historic gardens in European countries have a history of hundreds of years and have accumulated rich planning, design, and management experience. In comparison, modern landscape planning in China is relatively new and needs to accumulate more experience.

    \item \textbf{Technology and innovation}: Some developed countries have adopted advanced technologies and innovative methods in landscape planning. For example, they use advanced geographic information systems (GIS) \cite{marzouk2020planning,wu2024geospatial}, remote sensing technologies, and 3D visualization techniques to support planning and decision-making. In sustainable design, some countries actively apply concepts such as ecosystem services, low-impact development, and sustainable materials. China still has room for further development in these technologies and innovations.

    \item \textbf{Environmental awareness and sustainability}: Developed countries pay more attention to environmental awareness and sustainability in landscape planning. They focus on protecting natural ecosystems, reducing carbon emissions, and conserving energy and water resources. Some countries have formulated strict environmental regulations and standards, requiring planning and design to follow the principles of sustainable development. China has also made some efforts in this regard but still needs to further strengthen the practice of environmental awareness and sustainability.

    \item \textbf{Management and maintenance}: Developed countries emphasize the long-term management and maintenance of landscapes. They have established sound management mechanisms and teams to ensure the effective operation and maintenance of park facilities.  encompasses regular repairs, plant care, cleaning, safety, and so on. In China, the management and maintenance of some public parks still face challenges and need to be strengthened in terms of management capacity and investment.
\end{itemize}

Although there are gaps, China has also made significant progress in landscape planning and has unique advantages in some areas. The Chinese government and relevant departments have recognized these gaps and taken measures to strengthen the research, practice, and talent development of landscape planning. We believe that over time, the gap between China's landscape planning and the development abroad is expected to gradually narrow.

\section{Conclusion} \label{sec:conclusion} 

LA + AI provides new possibilities and innovative solutions to solve LA's current problems and challenges. By accelerating the design process, optimizing management and maintenance, enhancing ecological friendliness, and improving visitor experience, AI can bring greater benefits and sustainable development to LA. In this article, we first provide a comprehensive overview of various applications of AI technologies in LA, including their defining characteristics, key technologies, and various applications. Secondly, we discuss in detail how AI can assist the LA field in solving its current development problems, including urbanization, environmental degradation and ecological decline, irrational planning, insufficient management and maintenance, and lack of public participation. Finally, we elaborate on the relevant issues and challenges that need to be considered in the application of AI, to establish feasible solutions for AI to serve LA design. These include data acquisition and quality, the complexity and uncertainty of the landscape environment, human creativity and subjectivity, privacy, and security issues involving the collection and processing of personal and sensitive data, as well as the impact of user acceptance and education. The role of human professionals in the design, planning, and management process remains crucial for the future better utilization of AI's technological characteristics in LA.

\printcredits

\bibliographystyle{cas-model2-names}

\bibliography{main.bib}

\begin{thebibliography}{195}
\expandafter\ifx\csname natexlab\endcsname\relax\def\natexlab#1{#1}\fi
\providecommand{\url}[1]{\texttt{#1}}
\providecommand{\href}[2]{#2}
\providecommand{\path}[1]{#1}
\providecommand{\DOIprefix}{doi:}
\providecommand{\ArXivprefix}{arXiv:}
\providecommand{\URLprefix}{URL: }
\providecommand{\Pubmedprefix}{pmid:}
\providecommand{\doi}[1]{\href{http://dx.doi.org/#1}{\path{#1}}}
\providecommand{\Pubmed}[1]{\href{pmid:#1}{\path{#1}}}
\providecommand{\bibinfo}[2]{#2}
\ifx\xfnm\relax \def\xfnm[#1]{\unskip,\space#1}\fi
%Type = Article
\bibitem[{Abar et~al.(2017)Abar, Theodoropoulos, Lemarinier and O’Hare}]{abar2017agent}
\bibinfo{author}{Abar, S.}, \bibinfo{author}{Theodoropoulos, G.K.}, \bibinfo{author}{Lemarinier, P.}, \bibinfo{author}{O’Hare, G.M.}, \bibinfo{year}{2017}.
\newblock \bibinfo{title}{Agent based modelling and simulation tools: A review of the state-of-art software}.
\newblock \bibinfo{journal}{Computer Science Review} \bibinfo{volume}{24}, \bibinfo{pages}{13--33}.
%Type = Article
\bibitem[{Abdel-Hamid et~al.(2014)Abdel-Hamid, Mohamed, Jiang, Deng, Penn and Yu}]{abdel2014convolutional}
\bibinfo{author}{Abdel-Hamid, O.}, \bibinfo{author}{Mohamed, A.r.}, \bibinfo{author}{Jiang, H.}, \bibinfo{author}{Deng, L.}, \bibinfo{author}{Penn, G.}, \bibinfo{author}{Yu, D.}, \bibinfo{year}{2014}.
\newblock \bibinfo{title}{Convolutional neural networks for speech recognition}.
\newblock \bibinfo{journal}{IEEE/ACM Transactions on Audio, Speech, and Language Processing} \bibinfo{volume}{22}, \bibinfo{pages}{1533--1545}.
%Type = Article
\bibitem[{Abdi and Williams(2010)}]{abdi2010principal}
\bibinfo{author}{Abdi, H.}, \bibinfo{author}{Williams, L.J.}, \bibinfo{year}{2010}.
\newblock \bibinfo{title}{Principal component analysis}.
\newblock \bibinfo{journal}{Wiley Interdisciplinary Reviews: Computational Statistics} \bibinfo{volume}{2}, \bibinfo{pages}{433--459}.
%Type = Inproceedings
\bibitem[{Abed et~al.(2020)Abed, Al-Asfoor and Hussain}]{abed2020architectural}
\bibinfo{author}{Abed, M.H.}, \bibinfo{author}{Al-Asfoor, M.}, \bibinfo{author}{Hussain, Z.M.}, \bibinfo{year}{2020}.
\newblock \bibinfo{title}{Architectural heritage images classification using deep learning with {CNN}}, in: \bibinfo{booktitle}{The 2nd International Workshop on Visual Pattern Extraction and Recognition for Cultural Heritage Understandingco-located}, \bibinfo{publisher}{CEUR-WS Team}. pp. \bibinfo{pages}{1--12}.
%Type = Inproceedings
\bibitem[{Ahmad and Aliyu(2012)}]{ahmad2012need}
\bibinfo{author}{Ahmad, A.M.}, \bibinfo{author}{Aliyu, A.A.}, \bibinfo{year}{2012}.
\newblock \bibinfo{title}{The need for landscape information modelling ({LIM}) in landscape architecture}, in: \bibinfo{booktitle}{the 13th Digital Landscape Architecture Conference}, \bibinfo{organization}{Citeseer}. pp. \bibinfo{pages}{531--540}.
%Type = Book
\bibitem[{Alpaydin(2020)}]{alpaydin2020introduction}
\bibinfo{author}{Alpaydin, E.}, \bibinfo{year}{2020}.
\newblock \bibinfo{title}{Introduction to machine learning}.
\newblock \bibinfo{publisher}{MIT Press}.
%Type = Article
\bibitem[{Amabile(2020)}]{amabile2020creativity}
\bibinfo{author}{Amabile, T.M.}, \bibinfo{year}{2020}.
\newblock \bibinfo{title}{Creativity, artificial intelligence, and a world of surprises}.
\newblock \bibinfo{journal}{Academy of Management Discoveries} \bibinfo{volume}{6}, \bibinfo{pages}{351--354}.
%Type = Article
\bibitem[{Azhar(2011)}]{azhar2011building}
\bibinfo{author}{Azhar, S.}, \bibinfo{year}{2011}.
\newblock \bibinfo{title}{Building information modeling ({BIM}): Trends, benefits, risks, and challenges for the {AEC} industry}.
\newblock \bibinfo{journal}{Leadership and Management in Engineering} \bibinfo{volume}{11}, \bibinfo{pages}{241--252}.
%Type = Article
\bibitem[{Barcaui and Monat(2023)}]{barcaui2023better}
\bibinfo{author}{Barcaui, A.}, \bibinfo{author}{Monat, A.}, \bibinfo{year}{2023}.
\newblock \bibinfo{title}{Who is better in project planning? {G}enerative artificial intelligence or project managers?}
\newblock \bibinfo{journal}{Project Leadership and Society} \bibinfo{volume}{4}, \bibinfo{pages}{100101}.
%Type = Article
\bibitem[{Bartneck et~al.(2009)Bartneck, Kuli{\'c}, Croft and Zoghbi}]{bartneck2009measurement}
\bibinfo{author}{Bartneck, C.}, \bibinfo{author}{Kuli{\'c}, D.}, \bibinfo{author}{Croft, E.}, \bibinfo{author}{Zoghbi, S.}, \bibinfo{year}{2009}.
\newblock \bibinfo{title}{Measurement instruments for the anthropomorphism, animacy, likeability, perceived intelligence, and perceived safety of robots}.
\newblock \bibinfo{journal}{International Journal of Social Robotics} \bibinfo{volume}{1}, \bibinfo{pages}{71--81}.
%Type = Article
\bibitem[{Bentley et~al.(2014)Bentley, O'Brien and Brock}]{bentley2014mapping}
\bibinfo{author}{Bentley, R.A.}, \bibinfo{author}{O'Brien, M.J.}, \bibinfo{author}{Brock, W.A.}, \bibinfo{year}{2014}.
\newblock \bibinfo{title}{Mapping collective behavior in the big-data era}.
\newblock \bibinfo{journal}{Behavioral and Brain Sciences} \bibinfo{volume}{37}, \bibinfo{pages}{63--76}.
%Type = Inproceedings
\bibitem[{Beresford and Phillips(2000)}]{beresford2000protected}
\bibinfo{author}{Beresford, M.}, \bibinfo{author}{Phillips, A.}, \bibinfo{year}{2000}.
\newblock \bibinfo{title}{Protected landscapes: A conservation model for the 21st century}, in: \bibinfo{booktitle}{The George Wright Forum}, \bibinfo{organization}{JSTOR}. pp. \bibinfo{pages}{15--26}.
%Type = Article
\bibitem[{Bergen et~al.(1998)Bergen, McGaughey and Fridley}]{bergen1998data}
\bibinfo{author}{Bergen, S.D.}, \bibinfo{author}{McGaughey, R.J.}, \bibinfo{author}{Fridley, J.L.}, \bibinfo{year}{1998}.
\newblock \bibinfo{title}{Data-driven simulation, dimensional accuracy and realism in a landscape visualization tool}.
\newblock \bibinfo{journal}{Landscape and Urban Planning} \bibinfo{volume}{40}, \bibinfo{pages}{283--293}.
%Type = Article
\bibitem[{Bost et~al.(2014)Bost, Popa, Tu and Goldwasser}]{bost2014machine}
\bibinfo{author}{Bost, R.}, \bibinfo{author}{Popa, R.A.}, \bibinfo{author}{Tu, S.}, \bibinfo{author}{Goldwasser, S.}, \bibinfo{year}{2014}.
\newblock \bibinfo{title}{Machine learning classification over encrypted data}.
\newblock \bibinfo{journal}{Cryptology ePrint Archive} .
%Type = Article
\bibitem[{Bousquet and Le~Page(2004)}]{bousquet2004multi}
\bibinfo{author}{Bousquet, F.}, \bibinfo{author}{Le~Page, C.}, \bibinfo{year}{2004}.
\newblock \bibinfo{title}{Multi-agent simulations and ecosystem management: a review}.
\newblock \bibinfo{journal}{Ecological Modelling} \bibinfo{volume}{176}, \bibinfo{pages}{313--332}.
%Type = Article
\bibitem[{Breiman(2001)}]{breiman2001random}
\bibinfo{author}{Breiman, L.}, \bibinfo{year}{2001}.
\newblock \bibinfo{title}{Random forests}.
\newblock \bibinfo{journal}{Machine Learning} \bibinfo{volume}{45}, \bibinfo{pages}{5--32}.
%Type = Book
\bibitem[{Breiman(2017)}]{breiman2017classification}
\bibinfo{author}{Breiman, L.}, \bibinfo{year}{2017}.
\newblock \bibinfo{title}{Classification and regression trees}.
\newblock \bibinfo{publisher}{Routledge}.
%Type = Article
\bibitem[{Bryman(2006)}]{bryman2006integrating}
\bibinfo{author}{Bryman, A.}, \bibinfo{year}{2006}.
\newblock \bibinfo{title}{Integrating quantitative and qualitative research: how is it done?}
\newblock \bibinfo{journal}{Qualitative Rresearch} \bibinfo{volume}{6}, \bibinfo{pages}{97--113}.
%Type = Incollection
\bibitem[{Buckley et~al.(2015)Buckley, Cooke and Fayad}]{buckley2015using}
\bibinfo{author}{Buckley, K.}, \bibinfo{author}{Cooke, S.}, \bibinfo{author}{Fayad, S.}, \bibinfo{year}{2015}.
\newblock \bibinfo{title}{Using the historic urban landscape to re-imagine ballarat: The local context}, in: \bibinfo{booktitle}{Urban heritage, Development and Sustainability}. \bibinfo{publisher}{Routledge}, pp. \bibinfo{pages}{93--113}.
%Type = Inproceedings
\bibitem[{Buhalis and Amaranggana(2015)}]{buhalis2015smart}
\bibinfo{author}{Buhalis, D.}, \bibinfo{author}{Amaranggana, A.}, \bibinfo{year}{2015}.
\newblock \bibinfo{title}{Smart tourism destinations enhancing tourism experience through personalisation of services}, in: \bibinfo{booktitle}{Information and Communication Technologies in Tourism : the International Conference}, \bibinfo{organization}{Springer}. pp. \bibinfo{pages}{377--389}.
%Type = Article
\bibitem[{Buscombe and Ritchie(2018)}]{buscombe2018landscape}
\bibinfo{author}{Buscombe, D.}, \bibinfo{author}{Ritchie, A.C.}, \bibinfo{year}{2018}.
\newblock \bibinfo{title}{Landscape classification with deep neural networks}.
\newblock \bibinfo{journal}{Geosciences} \bibinfo{volume}{8}, \bibinfo{pages}{244}.
%Type = Article
\bibitem[{Bwambale et~al.(2022)Bwambale, Abagale and Anornu}]{bwambale2022smart}
\bibinfo{author}{Bwambale, E.}, \bibinfo{author}{Abagale, F.K.}, \bibinfo{author}{Anornu, G.K.}, \bibinfo{year}{2022}.
\newblock \bibinfo{title}{Smart irrigation monitoring and control strategies for improving water use efficiency in precision agriculture: A review}.
\newblock \bibinfo{journal}{Agricultural Water Management} \bibinfo{volume}{260}, \bibinfo{pages}{107324}.
%Type = Article
\bibitem[{Cai et~al.(2021)Cai, Fang, Zhang and Chen}]{cai2021joint}
\bibinfo{author}{Cai, Z.}, \bibinfo{author}{Fang, C.}, \bibinfo{author}{Zhang, Q.}, \bibinfo{author}{Chen, F.}, \bibinfo{year}{2021}.
\newblock \bibinfo{title}{Joint development of cultural heritage protection and tourism: the case of mount lushan cultural landscape heritage site}.
\newblock \bibinfo{journal}{Heritage Science} \bibinfo{volume}{9}, \bibinfo{pages}{86}.
%Type = Incollection
\bibitem[{Cantrell et~al.(2021)Cantrell, Zhang and Liu}]{cantrell2021artificial}
\bibinfo{author}{Cantrell, B.}, \bibinfo{author}{Zhang, Z.}, \bibinfo{author}{Liu, X.}, \bibinfo{year}{2021}.
\newblock \bibinfo{title}{Artificial intelligence and machine learning in landscape architecture}, in: \bibinfo{booktitle}{The Routledge Companion to Artificial Intelligence in Architecture}. \bibinfo{publisher}{Routledge}, pp. \bibinfo{pages}{232--247}.
%Type = Book
\bibitem[{Carr(1992)}]{carr1992public}
\bibinfo{author}{Carr, S.}, \bibinfo{year}{1992}.
\newblock \bibinfo{title}{Public space}.
\newblock \bibinfo{publisher}{Cambridge University Press}.
%Type = Article
\bibitem[{Carter and Grimwade(1997)}]{carter1997balancing}
\bibinfo{author}{Carter, B.}, \bibinfo{author}{Grimwade, G.}, \bibinfo{year}{1997}.
\newblock \bibinfo{title}{Balancing use and preservation in cultural heritage management}.
\newblock \bibinfo{journal}{International Journal of Heritage Studies} \bibinfo{volume}{3}, \bibinfo{pages}{45--53}.
%Type = Inproceedings
\bibitem[{Caruana and Niculescu-Mizil(2006)}]{caruana2006empirical}
\bibinfo{author}{Caruana, R.}, \bibinfo{author}{Niculescu-Mizil, A.}, \bibinfo{year}{2006}.
\newblock \bibinfo{title}{An empirical comparison of supervised learning algorithms}, in: \bibinfo{booktitle}{The 23rd International Conference on Machine Learning}, pp. \bibinfo{pages}{161--168}.
%Type = Article
\bibitem[{Chang and Lin(2011)}]{chang2011libsvm}
\bibinfo{author}{Chang, C.C.}, \bibinfo{author}{Lin, C.J.}, \bibinfo{year}{2011}.
\newblock \bibinfo{title}{{LIBSVM}: a library for support vector machines}.
\newblock \bibinfo{journal}{ACM Transactions on Intelligent Systems and Technology} \bibinfo{volume}{2}, \bibinfo{pages}{1--27}.
%Type = Article
\bibitem[{Chen et~al.(2020)Chen, Huang, Li, Chang and Huang}]{chen2020aiot}
\bibinfo{author}{Chen, C.J.}, \bibinfo{author}{Huang, Y.Y.}, \bibinfo{author}{Li, Y.S.}, \bibinfo{author}{Chang, C.Y.}, \bibinfo{author}{Huang, Y.M.}, \bibinfo{year}{2020}.
\newblock \bibinfo{title}{An {AI}o{T} based smart agricultural system for pests detection}.
\newblock \bibinfo{journal}{IEEE Access} \bibinfo{volume}{8}, \bibinfo{pages}{180750--180761}.
%Type = Article
\bibitem[{Chen et~al.(2019)Chen, Wang and Tong}]{chen2019aticnet}
\bibinfo{author}{Chen, J.}, \bibinfo{author}{Wang, C.}, \bibinfo{author}{Tong, Y.}, \bibinfo{year}{2019}.
\newblock \bibinfo{title}{Ati{CN}et: semantic segmentation with atrous spatial pyramid pooling in image cascade network}.
\newblock \bibinfo{journal}{Journal on Wireless Communications and Networking} \bibinfo{volume}{2019}, \bibinfo{pages}{1--7}.
%Type = Article
\bibitem[{Chen et~al.(2024a)Chen, Zhao, Yao, He, Li, Lian, Han, Yi and Li}]{chen2024enhancing}
\bibinfo{author}{Chen, R.}, \bibinfo{author}{Zhao, J.}, \bibinfo{author}{Yao, X.}, \bibinfo{author}{He, Y.}, \bibinfo{author}{Li, Y.}, \bibinfo{author}{Lian, Z.}, \bibinfo{author}{Han, Z.}, \bibinfo{author}{Yi, X.}, \bibinfo{author}{Li, H.}, \bibinfo{year}{2024}a.
\newblock \bibinfo{title}{Enhancing urban landscape design: A {GAN}-based approach for rapid color rendering of park sketches}.
\newblock \bibinfo{journal}{Land} \bibinfo{volume}{13}, \bibinfo{pages}{254}.
%Type = Inproceedings
\bibitem[{Chen et~al.(2024b)Chen, Gan, Sun, Wu and Yu}]{chen2024open}
\bibinfo{author}{Chen, Z.}, \bibinfo{author}{Gan, W.}, \bibinfo{author}{Sun, J.}, \bibinfo{author}{Wu, J.}, \bibinfo{author}{Yu, P.S.}, \bibinfo{year}{2024}b.
\newblock \bibinfo{title}{Open metaverse: Issues, evolution, and future}, in: \bibinfo{booktitle}{Companion Proceedings of the ACM on Web Conference}, pp. \bibinfo{pages}{1351--1360}.
%Type = Article
\bibitem[{Chen et~al.(2024c)Chen, Gan, Wu, Hu and Lin}]{chen2024data}
\bibinfo{author}{Chen, Z.}, \bibinfo{author}{Gan, W.}, \bibinfo{author}{Wu, J.}, \bibinfo{author}{Hu, K.}, \bibinfo{author}{Lin, H.}, \bibinfo{year}{2024}c.
\newblock \bibinfo{title}{Data scarcity in recommendation systems: A survey}.
\newblock \bibinfo{journal}{ACM Transactions on Recommender Systems} , \bibinfo{pages}{1--30}.
%Type = Inproceedings
\bibitem[{Chourabi et~al.(2012)Chourabi, Nam, Walker, Gil-Garcia, Mellouli, Nahon, Pardo and Scholl}]{chourabi2012understanding}
\bibinfo{author}{Chourabi, H.}, \bibinfo{author}{Nam, T.}, \bibinfo{author}{Walker, S.}, \bibinfo{author}{Gil-Garcia, J.R.}, \bibinfo{author}{Mellouli, S.}, \bibinfo{author}{Nahon, K.}, \bibinfo{author}{Pardo, T.A.}, \bibinfo{author}{Scholl, H.J.}, \bibinfo{year}{2012}.
\newblock \bibinfo{title}{Understanding smart cities: An integrative framework}, in: \bibinfo{booktitle}{the 45th Hawaii International Conference on System Sciences}, \bibinfo{organization}{IEEE}. pp. \bibinfo{pages}{2289--2297}.
%Type = Article
\bibitem[{Christin et~al.(2019)Christin, Hervet and Lecomte}]{christin2019applications}
\bibinfo{author}{Christin, S.}, \bibinfo{author}{Hervet, {\'E}.}, \bibinfo{author}{Lecomte, N.}, \bibinfo{year}{2019}.
\newblock \bibinfo{title}{Applications for deep learning in ecology}.
\newblock \bibinfo{journal}{Methods in Ecology and Evolution} \bibinfo{volume}{10}, \bibinfo{pages}{1632--1644}.
%Type = Article
\bibitem[{Collobert et~al.(2011)Collobert, Weston, Bottou, Karlen, Kavukcuoglu and Kuksa}]{collobert2011natural}
\bibinfo{author}{Collobert, R.}, \bibinfo{author}{Weston, J.}, \bibinfo{author}{Bottou, L.}, \bibinfo{author}{Karlen, M.}, \bibinfo{author}{Kavukcuoglu, K.}, \bibinfo{author}{Kuksa, P.}, \bibinfo{year}{2011}.
\newblock \bibinfo{title}{Natural language processing (almost) from scratch}.
\newblock \bibinfo{journal}{Journal of Machine Learning Research} \bibinfo{volume}{12}, \bibinfo{pages}{2493--2537}.
%Type = Article
\bibitem[{Cranz and Boland(2004)}]{cranz2004defining}
\bibinfo{author}{Cranz, G.}, \bibinfo{author}{Boland, M.}, \bibinfo{year}{2004}.
\newblock \bibinfo{title}{Defining the sustainable park: a fifth model for urban parks}.
\newblock \bibinfo{journal}{Landscape Journal} \bibinfo{volume}{23}, \bibinfo{pages}{102--120}.
%Type = Article
\bibitem[{Creswell et~al.(2018)Creswell, White, Dumoulin, Arulkumaran, Sengupta and Bharath}]{creswell2018generative}
\bibinfo{author}{Creswell, A.}, \bibinfo{author}{White, T.}, \bibinfo{author}{Dumoulin, V.}, \bibinfo{author}{Arulkumaran, K.}, \bibinfo{author}{Sengupta, B.}, \bibinfo{author}{Bharath, A.A.}, \bibinfo{year}{2018}.
\newblock \bibinfo{title}{Generative adversarial networks: An overview}.
\newblock \bibinfo{journal}{IEEE Signal Processing Magazine} \bibinfo{volume}{35}, \bibinfo{pages}{53--65}.
%Type = Article
\bibitem[{Cutler et~al.(2007)Cutler, Edwards~Jr, Beard, Cutler, Hess, Gibson and Lawler}]{cutler2007random}
\bibinfo{author}{Cutler, D.R.}, \bibinfo{author}{Edwards~Jr, T.C.}, \bibinfo{author}{Beard, K.H.}, \bibinfo{author}{Cutler, A.}, \bibinfo{author}{Hess, K.T.}, \bibinfo{author}{Gibson, J.}, \bibinfo{author}{Lawler, J.J.}, \bibinfo{year}{2007}.
\newblock \bibinfo{title}{Random forests for classification in ecology}.
\newblock \bibinfo{journal}{Ecology} \bibinfo{volume}{88}, \bibinfo{pages}{2783--2792}.
%Type = Book
\bibitem[{Dahlhaus and Thompson(2014)}]{dahlhaus2014visualising}
\bibinfo{author}{Dahlhaus, P.}, \bibinfo{author}{Thompson, H.}, \bibinfo{year}{2014}.
\newblock \bibinfo{title}{Visualising Ballarat-past, present, future}.
\newblock \bibinfo{publisher}{Federation University Australia}.
%Type = Article
\bibitem[{Dang et~al.(2020)Dang, Moreno-Garc{\'\i}a and De~la Prieta}]{dang2020sentiment}
\bibinfo{author}{Dang, N.C.}, \bibinfo{author}{Moreno-Garc{\'\i}a, M.N.}, \bibinfo{author}{De~la Prieta, F.}, \bibinfo{year}{2020}.
\newblock \bibinfo{title}{Sentiment analysis based on deep learning: A comparative study}.
\newblock \bibinfo{journal}{Electronics} \bibinfo{volume}{9}, \bibinfo{pages}{483}.
%Type = Article
\bibitem[{De~Paz et~al.(2016)De~Paz, Bajo, Rodr{\'\i}guez, Villarrubia and Corchado}]{de2016intelligent}
\bibinfo{author}{De~Paz, J.F.}, \bibinfo{author}{Bajo, J.}, \bibinfo{author}{Rodr{\'\i}guez, S.}, \bibinfo{author}{Villarrubia, G.}, \bibinfo{author}{Corchado, J.M.}, \bibinfo{year}{2016}.
\newblock \bibinfo{title}{Intelligent system for lighting control in smart cities}.
\newblock \bibinfo{journal}{Information Sciences} \bibinfo{volume}{372}, \bibinfo{pages}{241--255}.
%Type = Inproceedings
\bibitem[{Devlin et~al.(2019)Devlin, Chang, Lee and Toutanova}]{devlin2018bert}
\bibinfo{author}{Devlin, J.}, \bibinfo{author}{Chang, M.}, \bibinfo{author}{Lee, K.}, \bibinfo{author}{Toutanova, K.}, \bibinfo{year}{2019}.
\newblock \bibinfo{title}{{BERT}: Pre-training of deep bidirectional transformers for language understanding}, in: \bibinfo{booktitle}{The Conference of the North American Chapter of the Association for Computational Linguistics: Human Language Technologies}, pp. \bibinfo{pages}{4171--4186}.
%Type = Article
\bibitem[{Dorri et~al.(2018)Dorri, Kanhere and Jurdak}]{dorri2018multi}
\bibinfo{author}{Dorri, A.}, \bibinfo{author}{Kanhere, S.S.}, \bibinfo{author}{Jurdak, R.}, \bibinfo{year}{2018}.
\newblock \bibinfo{title}{Multi-agent systems: A survey}.
\newblock \bibinfo{journal}{IEEE Access} \bibinfo{volume}{6}, \bibinfo{pages}{28573--28593}.
%Type = Article
\bibitem[{Dou et~al.(2018)Dou, Qin, Jin and Li}]{dou2018knowledge}
\bibinfo{author}{Dou, J.}, \bibinfo{author}{Qin, J.}, \bibinfo{author}{Jin, Z.}, \bibinfo{author}{Li, Z.}, \bibinfo{year}{2018}.
\newblock \bibinfo{title}{Knowledge graph based on domain ontology and natural language processing technology for chinese intangible cultural heritage}.
\newblock \bibinfo{journal}{Journal of Visual Languages \& Computing} \bibinfo{volume}{48}, \bibinfo{pages}{19--28}.
%Type = Inproceedings
\bibitem[{Dubey et~al.(2016)Dubey, Naik, Parikh, Raskar and Hidalgo}]{dubey2016deep}
\bibinfo{author}{Dubey, A.}, \bibinfo{author}{Naik, N.}, \bibinfo{author}{Parikh, D.}, \bibinfo{author}{Raskar, R.}, \bibinfo{author}{Hidalgo, C.A.}, \bibinfo{year}{2016}.
\newblock \bibinfo{title}{Deep learning the city: Quantifying urban perception at a global scale}, in: \bibinfo{booktitle}{Computer Vision--ECCV}, \bibinfo{organization}{Springer}. pp. \bibinfo{pages}{196--212}.
%Type = Article
\bibitem[{Dy and Brodley(2004)}]{dy2004feature}
\bibinfo{author}{Dy, J.G.}, \bibinfo{author}{Brodley, C.E.}, \bibinfo{year}{2004}.
\newblock \bibinfo{title}{Feature selection for unsupervised learning}.
\newblock \bibinfo{journal}{Journal of Machine Learning Research} \bibinfo{volume}{5}, \bibinfo{pages}{845--889}.
%Type = Inproceedings
\bibitem[{Dzhusupova et~al.(2022)Dzhusupova, Bosch and Olsson}]{dzhusupova2022challenges}
\bibinfo{author}{Dzhusupova, R.}, \bibinfo{author}{Bosch, J.}, \bibinfo{author}{Olsson, H.H.}, \bibinfo{year}{2022}.
\newblock \bibinfo{title}{Challenges in developing and deploying ai in the engineering, procurement and construction industry}, in: \bibinfo{booktitle}{IEEE 46th Annual Computers, Software, and Applications Conference}, \bibinfo{organization}{IEEE}. pp. \bibinfo{pages}{1070--1075}.
%Type = Book
\bibitem[{Eastman et~al.(1974)Eastman, David, Gilles, Joseph, Douglas and Christos}]{eastman1974outline}
\bibinfo{author}{Eastman, C.}, \bibinfo{author}{David, F.}, \bibinfo{author}{Gilles, L.}, \bibinfo{author}{Joseph, L.}, \bibinfo{author}{Douglas, S.}, \bibinfo{author}{Christos, Y.}, \bibinfo{year}{1974}.
\newblock \bibinfo{title}{An Outline of the Building Description System. Research Report No. 50.}
\newblock \bibinfo{publisher}{Institute of Physical Planning, Carnegie-Mellon University}.
%Type = Article
\bibitem[{Elliott and Soifer(2022)}]{elliott2022ai}
\bibinfo{author}{Elliott, D.}, \bibinfo{author}{Soifer, E.}, \bibinfo{year}{2022}.
\newblock \bibinfo{title}{{AI} technologies, privacy, and security}.
\newblock \bibinfo{journal}{Frontiers in Artificial Intelligence} \bibinfo{volume}{5}, \bibinfo{pages}{826737}.
%Type = Article
\bibitem[{Esposito et~al.(2020)Esposito, Santoro and Camarda}]{esposito2020agent}
\bibinfo{author}{Esposito, D.}, \bibinfo{author}{Santoro, S.}, \bibinfo{author}{Camarda, D.}, \bibinfo{year}{2020}.
\newblock \bibinfo{title}{Agent-based analysis of urban spaces using space syntax and spatial cognition approaches: A case study in {B}ari, {I}taly}.
\newblock \bibinfo{journal}{Sustainability} \bibinfo{volume}{12}, \bibinfo{pages}{4625}.
%Type = Book
\bibitem[{Ferber and Weiss(1999)}]{ferber1999multi}
\bibinfo{author}{Ferber, J.}, \bibinfo{author}{Weiss, G.}, \bibinfo{year}{1999}.
\newblock \bibinfo{title}{Multi-agent systems: an introduction to distributed artificial intelligence}. volume~\bibinfo{volume}{1}.
\newblock \bibinfo{publisher}{Addison-wesley Reading}.
%Type = Article
\bibitem[{Ferentinos(2018)}]{ferentinos2018deep}
\bibinfo{author}{Ferentinos, K.P.}, \bibinfo{year}{2018}.
\newblock \bibinfo{title}{Deep learning models for plant disease detection and diagnosis}.
\newblock \bibinfo{journal}{Computers and Electronics in Agriculture} \bibinfo{volume}{145}, \bibinfo{pages}{311--318}.
%Type = Article
\bibitem[{Filor(1994)}]{filor1994nature}
\bibinfo{author}{Filor, S.W.}, \bibinfo{year}{1994}.
\newblock \bibinfo{title}{The nature of landscape design and design process}.
\newblock \bibinfo{journal}{Landscape and Urban Planning} \bibinfo{volume}{30}, \bibinfo{pages}{121--129}.
%Type = Article
\bibitem[{Francis(2001)}]{francis2001case}
\bibinfo{author}{Francis, M.}, \bibinfo{year}{2001}.
\newblock \bibinfo{title}{A case study method for landscape architecture}.
\newblock \bibinfo{journal}{Landscape Journal} \bibinfo{volume}{20}, \bibinfo{pages}{15--29}.
%Type = Article
\bibitem[{Fu(2023)}]{fu2023enhancing}
\bibinfo{author}{Fu, W.}, \bibinfo{year}{2023}.
\newblock \bibinfo{title}{Enhancing university campus landscape design through regression analysis: Integrating ecological environmental protection}.
\newblock \bibinfo{journal}{Soft Computing} \bibinfo{volume}{27}, \bibinfo{pages}{16309--16329}.
%Type = Misc
\bibitem[{Fui-Hoon~Nah et~al.(2023)Fui-Hoon~Nah, Zheng, Cai, Siau and Chen}]{fui2023generative}
\bibinfo{author}{Fui-Hoon~Nah, F.}, \bibinfo{author}{Zheng, R.}, \bibinfo{author}{Cai, J.}, \bibinfo{author}{Siau, K.}, \bibinfo{author}{Chen, L.}, \bibinfo{year}{2023}.
\newblock \bibinfo{title}{Generative {AI} and {C}hat{GPT}: Applications, challenges, and {AI}-human collaboration}.
%Type = Inproceedings
\bibitem[{Gabrilovich et~al.(2007)Gabrilovich, Markovitch et~al.}]{gabrilovich2007computing}
\bibinfo{author}{Gabrilovich, E.}, \bibinfo{author}{Markovitch, S.}, et~al., \bibinfo{year}{2007}.
\newblock \bibinfo{title}{Computing semantic relatedness using wikipedia-based explicit semantic analysis.}, in: \bibinfo{booktitle}{IJcAI}, pp. \bibinfo{pages}{1606--1611}.
%Type = Article
\bibitem[{Gallent and Shaw(2007)}]{gallent2007spatial}
\bibinfo{author}{Gallent, N.}, \bibinfo{author}{Shaw, D.}, \bibinfo{year}{2007}.
\newblock \bibinfo{title}{Spatial planning, area action plans and the rural-urban fringe}.
\newblock \bibinfo{journal}{Journal of Environmental Planning and Management} \bibinfo{volume}{50}, \bibinfo{pages}{617--638}.
%Type = Article
\bibitem[{Gan et~al.(2023)Gan, Chen, Wan, Chen and Chen}]{gan2023anomaly}
\bibinfo{author}{Gan, W.}, \bibinfo{author}{Chen, L.}, \bibinfo{author}{Wan, S.}, \bibinfo{author}{Chen, J.}, \bibinfo{author}{Chen, C.M.}, \bibinfo{year}{2023}.
\newblock \bibinfo{title}{Anomaly rule detection in sequence data}.
\newblock \bibinfo{journal}{IEEE Transactions on Knowledge and Data Engineering} \bibinfo{volume}{35}, \bibinfo{pages}{12095--12108}.
%Type = Article
\bibitem[{Gan et~al.(2017)Gan, Lin, Chao and Zhan}]{gan2017data}
\bibinfo{author}{Gan, W.}, \bibinfo{author}{Lin, J.C.W.}, \bibinfo{author}{Chao, H.C.}, \bibinfo{author}{Zhan, J.}, \bibinfo{year}{2017}.
\newblock \bibinfo{title}{Data mining in distributed environment: a survey}.
\newblock \bibinfo{journal}{Wiley Interdisciplinary Reviews: Data Mining and Knowledge Discovery} \bibinfo{volume}{7}, \bibinfo{pages}{e1216}.
%Type = Article
\bibitem[{Gan et~al.(2020)Gan, Lin, Fournier-Viger, Chao and Yu}]{gan2020huopm}
\bibinfo{author}{Gan, W.}, \bibinfo{author}{Lin, J.C.W.}, \bibinfo{author}{Fournier-Viger, P.}, \bibinfo{author}{Chao, H.C.}, \bibinfo{author}{Yu, P.S.}, \bibinfo{year}{2020}.
\newblock \bibinfo{title}{{HUOPM}: High-utility occupancy pattern mining}.
\newblock \bibinfo{journal}{IEEE Transactions on Cybernetics} \bibinfo{volume}{50}, \bibinfo{pages}{1195--1208}.
%Type = Article
\bibitem[{Gan et~al.(2021a)Gan, Lin, Zhang, Fournier-Viger, Chao and Yu}]{gan2021fast}
\bibinfo{author}{Gan, W.}, \bibinfo{author}{Lin, J.C.W.}, \bibinfo{author}{Zhang, J.}, \bibinfo{author}{Fournier-Viger, P.}, \bibinfo{author}{Chao, H.C.}, \bibinfo{author}{Yu, P.S.}, \bibinfo{year}{2021}a.
\newblock \bibinfo{title}{Fast utility mining on sequence data}.
\newblock \bibinfo{journal}{IEEE Transactions on Cybernetics} \bibinfo{volume}{51}, \bibinfo{pages}{487--500}.
%Type = Article
\bibitem[{Gan et~al.(2021b)Gan, Lin, Zhang, Yin, Fournier-Viger, Chao and Yu}]{gan2021utility}
\bibinfo{author}{Gan, W.}, \bibinfo{author}{Lin, J.C.W.}, \bibinfo{author}{Zhang, J.}, \bibinfo{author}{Yin, H.}, \bibinfo{author}{Fournier-Viger, P.}, \bibinfo{author}{Chao, H.C.}, \bibinfo{author}{Yu, P.S.}, \bibinfo{year}{2021}b.
\newblock \bibinfo{title}{Utility mining across multi-dimensional sequences}.
\newblock \bibinfo{journal}{ACM Transactions on Knowledge Discovery from Data} \bibinfo{volume}{15}, \bibinfo{pages}{1--24}.
%Type = Article
\bibitem[{Gazvoda(2002)}]{gazvoda2002characteristics}
\bibinfo{author}{Gazvoda, D.}, \bibinfo{year}{2002}.
\newblock \bibinfo{title}{Characteristics of modern landscape architecture and its education}.
\newblock \bibinfo{journal}{Landscape and Urban Planning} \bibinfo{volume}{60}, \bibinfo{pages}{117--133}.
%Type = Article
\bibitem[{Giones and Brem(2017)}]{giones2017digital}
\bibinfo{author}{Giones, F.}, \bibinfo{author}{Brem, A.}, \bibinfo{year}{2017}.
\newblock \bibinfo{title}{Digital technology entrepreneurship: A definition and research agenda}.
\newblock \bibinfo{journal}{Technology Innovation Management Review} \bibinfo{volume}{7}.
%Type = Article
\bibitem[{Goap et~al.(2018)Goap, Sharma, Shukla and Krishna}]{goap2018iot}
\bibinfo{author}{Goap, A.}, \bibinfo{author}{Sharma, D.}, \bibinfo{author}{Shukla, A.K.}, \bibinfo{author}{Krishna, C.R.}, \bibinfo{year}{2018}.
\newblock \bibinfo{title}{An {I}o{T} based smart irrigation management system using machine learning and open source technologies}.
\newblock \bibinfo{journal}{Computers and Electronics in Agriculture} \bibinfo{volume}{155}, \bibinfo{pages}{41--49}.
%Type = Article
\bibitem[{Goodfellow et~al.(2014)Goodfellow, Pouget-Abadie, Mirza, Xu, Warde-Farley, Ozair, Courville and Bengio}]{goodfellow2014generative}
\bibinfo{author}{Goodfellow, I.}, \bibinfo{author}{Pouget-Abadie, J.}, \bibinfo{author}{Mirza, M.}, \bibinfo{author}{Xu, B.}, \bibinfo{author}{Warde-Farley, D.}, \bibinfo{author}{Ozair, S.}, \bibinfo{author}{Courville, A.}, \bibinfo{author}{Bengio, Y.}, \bibinfo{year}{2014}.
\newblock \bibinfo{title}{Generative adversarial nets}.
\newblock \bibinfo{journal}{Advances in Neural Information Processing Systems} \bibinfo{volume}{27}.
%Type = Inproceedings
\bibitem[{Graves et~al.(2013)Graves, Mohamed and Hinton}]{graves2013speech}
\bibinfo{author}{Graves, A.}, \bibinfo{author}{Mohamed, A.r.}, \bibinfo{author}{Hinton, G.}, \bibinfo{year}{2013}.
\newblock \bibinfo{title}{Speech recognition with deep recurrent neural networks}, in: \bibinfo{booktitle}{IEEE International Conference on Acoustics, Speech and Signal Processing}, \bibinfo{organization}{IEEE}. pp. \bibinfo{pages}{6645--6649}.
%Type = Article
\bibitem[{Gretzel et~al.(2015)Gretzel, Sigala, Xiang and Koo}]{gretzel2015smart}
\bibinfo{author}{Gretzel, U.}, \bibinfo{author}{Sigala, M.}, \bibinfo{author}{Xiang, Z.}, \bibinfo{author}{Koo, C.}, \bibinfo{year}{2015}.
\newblock \bibinfo{title}{Smart tourism: foundations and developments}.
\newblock \bibinfo{journal}{Electronic Markets} \bibinfo{volume}{25}, \bibinfo{pages}{179--188}.
%Type = Article
\bibitem[{Gu et~al.(2018)Gu, Wang, Kuen, Ma, Shahroudy, Shuai, Liu, Wang, Wang, Cai et~al.}]{gu2018recent}
\bibinfo{author}{Gu, J.}, \bibinfo{author}{Wang, Z.}, \bibinfo{author}{Kuen, J.}, \bibinfo{author}{Ma, L.}, \bibinfo{author}{Shahroudy, A.}, \bibinfo{author}{Shuai, B.}, \bibinfo{author}{Liu, T.}, \bibinfo{author}{Wang, X.}, \bibinfo{author}{Wang, G.}, \bibinfo{author}{Cai, J.}, et~al., \bibinfo{year}{2018}.
\newblock \bibinfo{title}{Recent advances in convolutional neural networks}.
\newblock \bibinfo{journal}{Pattern Recognition} \bibinfo{volume}{77}, \bibinfo{pages}{354--377}.
%Type = Article
\bibitem[{Guilford(1967)}]{guilford1967nature}
\bibinfo{author}{Guilford, J.P.}, \bibinfo{year}{1967}.
\newblock \bibinfo{title}{The nature of human intelligence}.
\newblock \bibinfo{journal}{McGr awHill} \bibinfo{volume}{152}.
%Type = Article
\bibitem[{Guo et~al.(2015)Guo, Zhang and Zhu}]{guo2015earth}
\bibinfo{author}{Guo, H.D.}, \bibinfo{author}{Zhang, L.}, \bibinfo{author}{Zhu, L.W.}, \bibinfo{year}{2015}.
\newblock \bibinfo{title}{Earth observation big data for climate change research}.
\newblock \bibinfo{journal}{Advances in Climate Change Research} \bibinfo{volume}{6}, \bibinfo{pages}{108--117}.
%Type = Article
\bibitem[{Gupta and Bhadauria(2014)}]{gupta2014object}
\bibinfo{author}{Gupta, N.}, \bibinfo{author}{Bhadauria, H.}, \bibinfo{year}{2014}.
\newblock \bibinfo{title}{Object based information extraction from high resolution satellite imagery using ecognition}.
\newblock \bibinfo{journal}{International Journal of Computer Science Issues} \bibinfo{volume}{11}, \bibinfo{pages}{139--144}.
%Type = Article
\bibitem[{Hashem et~al.(2016)Hashem, Chang, Anuar, Adewole, Yaqoob, Gani, Ahmed and Chiroma}]{hashem2016role}
\bibinfo{author}{Hashem, I.A.T.}, \bibinfo{author}{Chang, V.}, \bibinfo{author}{Anuar, N.B.}, \bibinfo{author}{Adewole, K.}, \bibinfo{author}{Yaqoob, I.}, \bibinfo{author}{Gani, A.}, \bibinfo{author}{Ahmed, E.}, \bibinfo{author}{Chiroma, H.}, \bibinfo{year}{2016}.
\newblock \bibinfo{title}{The role of big data in smart city}.
\newblock \bibinfo{journal}{International Journal of Information Management} \bibinfo{volume}{36}, \bibinfo{pages}{748--758}.
%Type = Inproceedings
\bibitem[{He et~al.(2016)He, Zhang, Ren and Sun}]{he2016deep}
\bibinfo{author}{He, K.}, \bibinfo{author}{Zhang, X.}, \bibinfo{author}{Ren, S.}, \bibinfo{author}{Sun, J.}, \bibinfo{year}{2016}.
\newblock \bibinfo{title}{Deep residual learning for image recognition}, in: \bibinfo{booktitle}{The IEEE Conference on Computer Vision and Pattern Recognition}, pp. \bibinfo{pages}{770--778}.
%Type = Article
\bibitem[{Herr and Kvan(2007)}]{herr2007adapting}
\bibinfo{author}{Herr, C.M.}, \bibinfo{author}{Kvan, T.}, \bibinfo{year}{2007}.
\newblock \bibinfo{title}{Adapting cellular automata to support the architectural design process}.
\newblock \bibinfo{journal}{Automation in Construction} \bibinfo{volume}{16}, \bibinfo{pages}{61--69}.
%Type = Book
\bibitem[{Hillier(2007)}]{hillier2007space}
\bibinfo{author}{Hillier, B.}, \bibinfo{year}{2007}.
\newblock \bibinfo{title}{Space is the machine: a configurational theory of architecture}.
\newblock \bibinfo{publisher}{Space Syntax}.
%Type = Article
\bibitem[{Hochreiter and Schmidhuber(1997)}]{hochreiter1997long}
\bibinfo{author}{Hochreiter, S.}, \bibinfo{author}{Schmidhuber, J.}, \bibinfo{year}{1997}.
\newblock \bibinfo{title}{Long short-term memory}.
\newblock \bibinfo{journal}{Neural Computation} \bibinfo{volume}{9}, \bibinfo{pages}{1735--1780}.
%Type = Book
\bibitem[{Hosmer~Jr et~al.(2013)Hosmer~Jr, Lemeshow and Sturdivant}]{hosmer2013applied}
\bibinfo{author}{Hosmer~Jr, D.W.}, \bibinfo{author}{Lemeshow, S.}, \bibinfo{author}{Sturdivant, R.X.}, \bibinfo{year}{2013}.
\newblock \bibinfo{title}{Applied logistic regression}.
\newblock \bibinfo{publisher}{John Wiley \& Sons}.
%Type = Article
\bibitem[{Hu et~al.(2021)Hu, Lu, Pan, Gong and Yang}]{hu2021can}
\bibinfo{author}{Hu, Q.}, \bibinfo{author}{Lu, Y.}, \bibinfo{author}{Pan, Z.}, \bibinfo{author}{Gong, Y.}, \bibinfo{author}{Yang, Z.}, \bibinfo{year}{2021}.
\newblock \bibinfo{title}{Can {AI} artifacts influence human cognition? the effects of artificial autonomy in intelligent personal assistants}.
\newblock \bibinfo{journal}{International Journal of Information Management} \bibinfo{volume}{56}, \bibinfo{pages}{102250}.
%Type = Article
\bibitem[{Huai and Van~de Voorde(2022)}]{huai2022environmental}
\bibinfo{author}{Huai, S.}, \bibinfo{author}{Van~de Voorde, T.}, \bibinfo{year}{2022}.
\newblock \bibinfo{title}{Which environmental features contribute to positive and negative perceptions of urban parks? a cross-cultural comparison using online reviews and natural language processing methods}.
\newblock \bibinfo{journal}{Landscape and Urban Planning} \bibinfo{volume}{218}, \bibinfo{pages}{104307}.
%Type = Article
\bibitem[{Huang et~al.(2024)Huang, Gan and Yu}]{huang2024taspm}
\bibinfo{author}{Huang, G.}, \bibinfo{author}{Gan, W.}, \bibinfo{author}{Yu, P.S.}, \bibinfo{year}{2024}.
\newblock \bibinfo{title}{{TaSPM}: Targeted sequential pattern mining}.
\newblock \bibinfo{journal}{ACM Transactions on Knowledge Discovery from Data} \bibinfo{volume}{18}, \bibinfo{pages}{1--18}.
%Type = Inproceedings
\bibitem[{Huang and Zheng(2018)}]{huang2018architectural}
\bibinfo{author}{Huang, W.}, \bibinfo{author}{Zheng, H.}, \bibinfo{year}{2018}.
\newblock \bibinfo{title}{Architectural drawings recognition and generation through machine learning}, in: \bibinfo{booktitle}{The 38th Annual Conference of the Association for Computer Aided Design in Architecture}, pp. \bibinfo{pages}{18--20}.
%Type = Article
\bibitem[{Ibrahim et~al.(2023)Ibrahim, Khattab, Khattab and Abraham}]{ibrahim2023expatriates}
\bibinfo{author}{Ibrahim, H.}, \bibinfo{author}{Khattab, Z.}, \bibinfo{author}{Khattab, T.}, \bibinfo{author}{Abraham, R.}, \bibinfo{year}{2023}.
\newblock \bibinfo{title}{Expatriates’ housing dispersal outlook in a rapidly developing metropolis based on urban growth predicted using a machine learning algorithm}.
\newblock \bibinfo{journal}{Housing Policy Debate} \bibinfo{volume}{33}, \bibinfo{pages}{641--661}.
%Type = Article
\bibitem[{Inskeep(1987)}]{inskeep1987environmental}
\bibinfo{author}{Inskeep, E.}, \bibinfo{year}{1987}.
\newblock \bibinfo{title}{Environmental planning for tourism}.
\newblock \bibinfo{journal}{Annals of Tourism Research} \bibinfo{volume}{14}, \bibinfo{pages}{118--135}.
%Type = Inproceedings
\bibitem[{Isola et~al.(2017)Isola, Zhu, Zhou and Efros}]{isola2017image}
\bibinfo{author}{Isola, P.}, \bibinfo{author}{Zhu, J.}, \bibinfo{author}{Zhou, T.}, \bibinfo{author}{Efros, A.A.}, \bibinfo{year}{2017}.
\newblock \bibinfo{title}{Image-to-image translation with conditional adversarial networks}, in: \bibinfo{booktitle}{The IEEE Conference on Computer Vision and Pattern Recognition}, pp. \bibinfo{pages}{1125--1134}.
%Type = Article
\bibitem[{Jain et~al.(1996)Jain, Mao and Mohiuddin}]{jain1996artificial}
\bibinfo{author}{Jain, A.K.}, \bibinfo{author}{Mao, J.}, \bibinfo{author}{Mohiuddin, K.M.}, \bibinfo{year}{1996}.
\newblock \bibinfo{title}{Artificial neural networks: A tutorial}.
\newblock \bibinfo{journal}{Computer} \bibinfo{volume}{29}, \bibinfo{pages}{31--44}.
%Type = Book
\bibitem[{Johnson(2002)}]{johnson2002emergence}
\bibinfo{author}{Johnson, S.}, \bibinfo{year}{2002}.
\newblock \bibinfo{title}{Emergence: The connected lives of ants, brains, cities, and software}.
\newblock \bibinfo{publisher}{Simon and Schuster}.
%Type = Article
\bibitem[{Jordan and Mitchell(2015)}]{jordan2015machine}
\bibinfo{author}{Jordan, M.I.}, \bibinfo{author}{Mitchell, T.M.}, \bibinfo{year}{2015}.
\newblock \bibinfo{title}{Machine learning: Trends, perspectives, and prospects}.
\newblock \bibinfo{journal}{Science} \bibinfo{volume}{349}, \bibinfo{pages}{255--260}.
%Type = Incollection
\bibitem[{Kennedy(2006)}]{kennedy2006swarm}
\bibinfo{author}{Kennedy, J.}, \bibinfo{year}{2006}.
\newblock \bibinfo{title}{Swarm intelligence}, in: \bibinfo{booktitle}{Handbook of nature-inspired and innovative computing: integrating classical models with emerging technologies}. \bibinfo{publisher}{Springer}, pp. \bibinfo{pages}{187--219}.
%Type = Article
\bibitem[{Kim et~al.(2016)Kim, Walewski and Cho}]{kim2016multiobjective}
\bibinfo{author}{Kim, K.}, \bibinfo{author}{Walewski, J.}, \bibinfo{author}{Cho, Y.K.}, \bibinfo{year}{2016}.
\newblock \bibinfo{title}{Multiobjective construction schedule optimization using modified niched pareto genetic algorithm}.
\newblock \bibinfo{journal}{Journal of Management in Engineering} \bibinfo{volume}{32}, \bibinfo{pages}{04015038}.
%Type = Book
\bibitem[{Koehn(2009)}]{koehn2009statistical}
\bibinfo{author}{Koehn, P.}, \bibinfo{year}{2009}.
\newblock \bibinfo{title}{Statistical machine translation}.
\newblock \bibinfo{publisher}{Cambridge University Press}.
%Type = Article
\bibitem[{Korteling et~al.(2021)Korteling, van~de Boer-Visschedijk, Blankendaal, Boonekamp and Eikelboom}]{korteling2021human}
\bibinfo{author}{Korteling, J.H.}, \bibinfo{author}{van~de Boer-Visschedijk, G.C.}, \bibinfo{author}{Blankendaal, R.A.}, \bibinfo{author}{Boonekamp, R.C.}, \bibinfo{author}{Eikelboom, A.R.}, \bibinfo{year}{2021}.
\newblock \bibinfo{title}{Human-versus artificial intelligence}.
\newblock \bibinfo{journal}{Frontiers in Artificial Intelligence} \bibinfo{volume}{4}, \bibinfo{pages}{622364}.
%Type = Article
\bibitem[{Kumar and Cheng(2015)}]{kumar2015bim}
\bibinfo{author}{Kumar, S.S.}, \bibinfo{author}{Cheng, J.C.}, \bibinfo{year}{2015}.
\newblock \bibinfo{title}{A {BIM}-based automated site layout planning framework for congested construction sites}.
\newblock \bibinfo{journal}{Automation in Construction} \bibinfo{volume}{59}, \bibinfo{pages}{24--37}.
%Type = Inproceedings
\bibitem[{Lafferty et~al.(2001)Lafferty, McCallum, Pereira et~al.}]{lafferty2001conditional}
\bibinfo{author}{Lafferty, J.}, \bibinfo{author}{McCallum, A.}, \bibinfo{author}{Pereira, F.}, et~al., \bibinfo{year}{2001}.
\newblock \bibinfo{title}{Conditional random fields: Probabilistic models for segmenting and labeling sequence data}, in: \bibinfo{booktitle}{ICML}, \bibinfo{organization}{Williamstown, MA}. p.~\bibinfo{pages}{3}.
%Type = Article
\bibitem[{LeCun et~al.(2015)LeCun, Bengio and Hinton}]{lecun2015deep}
\bibinfo{author}{LeCun, Y.}, \bibinfo{author}{Bengio, Y.}, \bibinfo{author}{Hinton, G.}, \bibinfo{year}{2015}.
\newblock \bibinfo{title}{Deep learning}.
\newblock \bibinfo{journal}{Nature} \bibinfo{volume}{521}, \bibinfo{pages}{436--444}.
%Type = Article
\bibitem[{Lee et~al.(2024)Lee, Law and Hoffman}]{lee2024when}
\bibinfo{author}{Lee, S.y.}, \bibinfo{author}{Law, M.}, \bibinfo{author}{Hoffman, G.}, \bibinfo{year}{2024}.
\newblock \bibinfo{title}{When and how to use {AI} in the design process? {I}mplications for human-{AI} design collaboration}.
\newblock \bibinfo{journal}{International Journal of Human--Computer Interaction} , \bibinfo{pages}{1--16}.
%Type = Article
\bibitem[{Leitao and Ahern(2002)}]{leitao2002applying}
\bibinfo{author}{Leitao, A.B.}, \bibinfo{author}{Ahern, J.}, \bibinfo{year}{2002}.
\newblock \bibinfo{title}{Applying landscape ecological concepts and metrics in sustainable landscape planning}.
\newblock \bibinfo{journal}{Landscape and Urban Planning} \bibinfo{volume}{59}, \bibinfo{pages}{65--93}.
%Type = Article
\bibitem[{Levinthal and Warglien(1999)}]{levinthal1999landscape}
\bibinfo{author}{Levinthal, D.A.}, \bibinfo{author}{Warglien, M.}, \bibinfo{year}{1999}.
\newblock \bibinfo{title}{Landscape design: Designing for local action in complex worlds}.
\newblock \bibinfo{journal}{Organization Science} \bibinfo{volume}{10}, \bibinfo{pages}{342--357}.
%Type = Article
\bibitem[{Li et~al.(2021)Li, CAO, He, He, Cao, Wang and Fang}]{li2021understanding}
\bibinfo{author}{Li, G.}, \bibinfo{author}{CAO, Y.}, \bibinfo{author}{He, Z.}, \bibinfo{author}{He, J.}, \bibinfo{author}{Cao, Y.}, \bibinfo{author}{Wang, J.}, \bibinfo{author}{Fang, X.}, \bibinfo{year}{2021}.
\newblock \bibinfo{title}{Understanding the diversity of urban–rural fringe development in a fast urbanizing region of china}.
\newblock \bibinfo{journal}{Remote Sensing} \bibinfo{volume}{13}.
%Type = Article
\bibitem[{Li and Hsu(2020)}]{li2020automated}
\bibinfo{author}{Li, W.}, \bibinfo{author}{Hsu, C.Y.}, \bibinfo{year}{2020}.
\newblock \bibinfo{title}{Automated terrain feature identification from remote sensing imagery: a deep learning approach}.
\newblock \bibinfo{journal}{International Journal of Geographical Information Science} \bibinfo{volume}{34}, \bibinfo{pages}{637--660}.
%Type = Inproceedings
\bibitem[{Lin et~al.(2015)Lin, Gan, Fournier-Viger and Hong}]{lin2015mining}
\bibinfo{author}{Lin, J.C.W.}, \bibinfo{author}{Gan, W.}, \bibinfo{author}{Fournier-Viger, P.}, \bibinfo{author}{Hong, T.P.}, \bibinfo{year}{2015}.
\newblock \bibinfo{title}{Mining high-utility itemsets with multiple minimum utility thresholds}, in: \bibinfo{booktitle}{The Eighth International C* Conference on Computer Science \& Software Engineering}, pp. \bibinfo{pages}{9--17}.
%Type = Article
\bibitem[{Liu et~al.(2020a)Liu, Cao, Yang, Zhou and Ai}]{liu2020pattern}
\bibinfo{author}{Liu, C.}, \bibinfo{author}{Cao, Y.}, \bibinfo{author}{Yang, C.}, \bibinfo{author}{Zhou, Y.}, \bibinfo{author}{Ai, M.}, \bibinfo{year}{2020}a.
\newblock \bibinfo{title}{Pattern identification and analysis for the traditional village using low altitude {UAV}-borne remote sensing: Multifeatured geospatial data to support rural landscape investigation, documentation and management}.
\newblock \bibinfo{journal}{Journal of Cultural Heritage} \bibinfo{volume}{44}, \bibinfo{pages}{185--195}.
%Type = Article
\bibitem[{Liu et~al.(2021)Liu, Fang, Dong and Xu}]{liu2021review}
\bibinfo{author}{Liu, M.}, \bibinfo{author}{Fang, S.}, \bibinfo{author}{Dong, H.}, \bibinfo{author}{Xu, C.}, \bibinfo{year}{2021}.
\newblock \bibinfo{title}{Review of digital twin about concepts, technologies, and industrial applications}.
\newblock \bibinfo{journal}{Journal of Manufacturing Systems} \bibinfo{volume}{58}, \bibinfo{pages}{346--361}.
%Type = Article
\bibitem[{Liu et~al.(2014)Liu, Ma, Li, Ai, Li and He}]{liu2014simulating}
\bibinfo{author}{Liu, X.}, \bibinfo{author}{Ma, L.}, \bibinfo{author}{Li, X.}, \bibinfo{author}{Ai, B.}, \bibinfo{author}{Li, S.}, \bibinfo{author}{He, Z.}, \bibinfo{year}{2014}.
\newblock \bibinfo{title}{Simulating urban growth by integrating landscape expansion index ({LEI}) and cellular automata}.
\newblock \bibinfo{journal}{International Journal of Geographical Information Science} \bibinfo{volume}{28}, \bibinfo{pages}{148--163}.
%Type = Article
\bibitem[{Liu et~al.(2020b)Liu, Xie, Wang, Zou, Xiong, Ying and Vasilakos}]{liu2020privacy}
\bibinfo{author}{Liu, X.}, \bibinfo{author}{Xie, L.}, \bibinfo{author}{Wang, Y.}, \bibinfo{author}{Zou, J.}, \bibinfo{author}{Xiong, J.}, \bibinfo{author}{Ying, Z.}, \bibinfo{author}{Vasilakos, A.V.}, \bibinfo{year}{2020}b.
\newblock \bibinfo{title}{Privacy and security issues in deep learning: A survey}.
\newblock \bibinfo{journal}{IEEE Access} \bibinfo{volume}{9}, \bibinfo{pages}{4566--4593}.
%Type = Article
\bibitem[{Liu et~al.(2017)Liu, Van~Nederveen and Hertogh}]{liu2017understanding}
\bibinfo{author}{Liu, Y.}, \bibinfo{author}{Van~Nederveen, S.}, \bibinfo{author}{Hertogh, M.}, \bibinfo{year}{2017}.
\newblock \bibinfo{title}{Understanding effects of {BIM} on collaborative design and construction: An empirical study in china}.
\newblock \bibinfo{journal}{International Journal of Project Management} \bibinfo{volume}{35}, \bibinfo{pages}{686--698}.
%Type = Article
\bibitem[{Lund et~al.(2017)Lund, {\O}stergaard, Connolly and Mathiesen}]{lund2017smart}
\bibinfo{author}{Lund, H.}, \bibinfo{author}{{\O}stergaard, P.A.}, \bibinfo{author}{Connolly, D.}, \bibinfo{author}{Mathiesen, B.V.}, \bibinfo{year}{2017}.
\newblock \bibinfo{title}{Smart energy and smart energy systems}.
\newblock \bibinfo{journal}{Energy} \bibinfo{volume}{137}, \bibinfo{pages}{556--565}.
%Type = Article
\bibitem[{Luo(2021)}]{luo2021online}
\bibinfo{author}{Luo, J.}, \bibinfo{year}{2021}.
\newblock \bibinfo{title}{Online design of green urban garden landscape based on machine learning and computer simulation technology}.
\newblock \bibinfo{journal}{Environmental Technology \& Innovation} \bibinfo{volume}{24}, \bibinfo{pages}{101819}.
%Type = Article
\bibitem[{Mair et~al.(2000)Mair, Kadoda, Lefley, Phalp, Schofield, Shepperd and Webster}]{mair2000investigation}
\bibinfo{author}{Mair, C.}, \bibinfo{author}{Kadoda, G.}, \bibinfo{author}{Lefley, M.}, \bibinfo{author}{Phalp, K.}, \bibinfo{author}{Schofield, C.}, \bibinfo{author}{Shepperd, M.}, \bibinfo{author}{Webster, S.}, \bibinfo{year}{2000}.
\newblock \bibinfo{title}{An investigation of machine learning based prediction systems}.
\newblock \bibinfo{journal}{Journal of systems and software} \bibinfo{volume}{53}, \bibinfo{pages}{23--29}.
%Type = Book
\bibitem[{Marsh(2005)}]{marsh2005landscape}
\bibinfo{author}{Marsh, W.M.}, \bibinfo{year}{2005}.
\newblock \bibinfo{title}{Landscape planning: Environmental applications}. volume~\bibinfo{volume}{4}.
\newblock \bibinfo{publisher}{Wiley New York}.
%Type = Article
\bibitem[{Marzouk and Othman(2020)}]{marzouk2020planning}
\bibinfo{author}{Marzouk, M.}, \bibinfo{author}{Othman, A.}, \bibinfo{year}{2020}.
\newblock \bibinfo{title}{Planning utility infrastructure requirements for smart cities using the integration between {BIM} and {GIS}}.
\newblock \bibinfo{journal}{Sustainable Cities and Society} \bibinfo{volume}{57}, \bibinfo{pages}{102120}.
%Type = Book
\bibitem[{Milani(2009)}]{milani2009art}
\bibinfo{author}{Milani, R.}, \bibinfo{year}{2009}.
\newblock \bibinfo{title}{Art of the Landscape}.
\newblock \bibinfo{publisher}{McGill-Queen's Press-MQUP}.
%Type = Article
\bibitem[{Minaee et~al.(2021)Minaee, Boykov, Porikli, Plaza, Kehtarnavaz and Terzopoulos}]{minaee2021image}
\bibinfo{author}{Minaee, S.}, \bibinfo{author}{Boykov, Y.}, \bibinfo{author}{Porikli, F.}, \bibinfo{author}{Plaza, A.}, \bibinfo{author}{Kehtarnavaz, N.}, \bibinfo{author}{Terzopoulos, D.}, \bibinfo{year}{2021}.
\newblock \bibinfo{title}{Image segmentation using deep learning: A survey}.
\newblock \bibinfo{journal}{IEEE Transactions on Pattern Analysis and Machine Intelligence} \bibinfo{volume}{44}, \bibinfo{pages}{3523--3542}.
%Type = Book
\bibitem[{Motloch(2000)}]{motloch2000introduction}
\bibinfo{author}{Motloch, J.L.}, \bibinfo{year}{2000}.
\newblock \bibinfo{title}{Introduction to landscape design}.
\newblock \bibinfo{publisher}{John Wiley \& Sons}.
%Type = Inproceedings
\bibitem[{Nasukawa and Yi(2003)}]{nasukawa2003sentiment}
\bibinfo{author}{Nasukawa, T.}, \bibinfo{author}{Yi, J.}, \bibinfo{year}{2003}.
\newblock \bibinfo{title}{Sentiment analysis: Capturing favorability using natural language processing}, in: \bibinfo{booktitle}{The 2nd International Conference on Knowledge Capture}, pp. \bibinfo{pages}{70--77}.
%Type = Book
\bibitem[{Newton(1971)}]{newton1971design}
\bibinfo{author}{Newton, N.T.}, \bibinfo{year}{1971}.
\newblock \bibinfo{title}{Design on the land: The development of landscape architecture}.
\newblock \bibinfo{publisher}{La Editorial, UPR}.
%Type = Book
\bibitem[{Nilsson(1971)}]{nilsson1971problem}
\bibinfo{author}{Nilsson, N.J.}, \bibinfo{year}{1971}.
\newblock \bibinfo{title}{Problem-Solving Methods in Artificial Intelligence}.
\newblock \bibinfo{publisher}{McGraw-Hill Pub. Co.}
%Type = Book
\bibitem[{Nilsson(1982)}]{nilsson1982principles}
\bibinfo{author}{Nilsson, N.J.}, \bibinfo{year}{1982}.
\newblock \bibinfo{title}{Principles of artificial intelligence}.
\newblock \bibinfo{publisher}{Morgan Kaufmann}.
%Type = Article
\bibitem[{Nishant et~al.(2020)Nishant, Kennedy and Corbett}]{nishant2020artificial}
\bibinfo{author}{Nishant, R.}, \bibinfo{author}{Kennedy, M.}, \bibinfo{author}{Corbett, J.}, \bibinfo{year}{2020}.
\newblock \bibinfo{title}{Artificial intelligence for sustainability: Challenges, opportunities, and a research agenda}.
\newblock \bibinfo{journal}{International Journal of Information Management} \bibinfo{volume}{53}, \bibinfo{pages}{102104}.
%Type = Article
\bibitem[{Oh(2001)}]{oh2001landscape}
\bibinfo{author}{Oh, K.}, \bibinfo{year}{2001}.
\newblock \bibinfo{title}{Landscape information system: A {GIS} approach to managing urban development}.
\newblock \bibinfo{journal}{Landscape and Urban Planning} \bibinfo{volume}{54}, \bibinfo{pages}{81--91}.
%Type = Article
\bibitem[{Otter et~al.(2020)Otter, Medina and Kalita}]{otter2020survey}
\bibinfo{author}{Otter, D.W.}, \bibinfo{author}{Medina, J.R.}, \bibinfo{author}{Kalita, J.K.}, \bibinfo{year}{2020}.
\newblock \bibinfo{title}{A survey of the usages of deep learning for natural language processing}.
\newblock \bibinfo{journal}{IEEE Transactions on Neural Networks and Learning Systems} \bibinfo{volume}{32}, \bibinfo{pages}{604--624}.
%Type = Article
\bibitem[{Papadimitriou(2012)}]{papadimitriou2012artificial}
\bibinfo{author}{Papadimitriou, F.}, \bibinfo{year}{2012}.
\newblock \bibinfo{title}{Artificial intelligence in modelling the complexity of mediterranean landscape transformations}.
\newblock \bibinfo{journal}{Computers and Electronics in Agriculture} \bibinfo{volume}{81}, \bibinfo{pages}{87--96}.
%Type = Article
\bibitem[{Pena et~al.(2021)Pena, Carballal, Rodr{\'\i}guez-Fern{\'a}ndez, Santos and Romero}]{pena2021artificial}
\bibinfo{author}{Pena, M.L.C.}, \bibinfo{author}{Carballal, A.}, \bibinfo{author}{Rodr{\'\i}guez-Fern{\'a}ndez, N.}, \bibinfo{author}{Santos, I.}, \bibinfo{author}{Romero, J.}, \bibinfo{year}{2021}.
\newblock \bibinfo{title}{Artificial intelligence applied to conceptual design. a review of its use in architecture}.
\newblock \bibinfo{journal}{Automation in Construction} \bibinfo{volume}{124}, \bibinfo{pages}{103550}.
%Type = Article
\bibitem[{Portman et~al.(2015)Portman, Natapov and Fisher-Gewirtzman}]{portman2015go}
\bibinfo{author}{Portman, M.E.}, \bibinfo{author}{Natapov, A.}, \bibinfo{author}{Fisher-Gewirtzman, D.}, \bibinfo{year}{2015}.
\newblock \bibinfo{title}{To go where no man has gone before: Virtual reality in architecture, landscape architecture and environmental planning}.
\newblock \bibinfo{journal}{Computers, Environment and Urban Systems} \bibinfo{volume}{54}, \bibinfo{pages}{376--384}.
%Type = Article
\bibitem[{Quinlan(1986)}]{quinlan1986induction}
\bibinfo{author}{Quinlan, J.R.}, \bibinfo{year}{1986}.
\newblock \bibinfo{title}{Induction of decision trees}.
\newblock \bibinfo{journal}{Machine Learning} \bibinfo{volume}{1}, \bibinfo{pages}{81--106}.
%Type = Article
\bibitem[{Raman and Naderi(2006)}]{raman2006computer}
\bibinfo{author}{Raman, B.}, \bibinfo{author}{Naderi, J.R.}, \bibinfo{year}{2006}.
\newblock \bibinfo{title}{Computer based pedestrian landscape design using decision tree templates}.
\newblock \bibinfo{journal}{Advanced Engineering Informatics} \bibinfo{volume}{20}, \bibinfo{pages}{23--30}.
%Type = Article
\bibitem[{Raum et~al.(2019)Raum, Hand, Hall, Edwards, O'brien and Doick}]{raum2019achieving}
\bibinfo{author}{Raum, S.}, \bibinfo{author}{Hand, K.}, \bibinfo{author}{Hall, C.}, \bibinfo{author}{Edwards, D.}, \bibinfo{author}{O'brien, L.}, \bibinfo{author}{Doick, K.}, \bibinfo{year}{2019}.
\newblock \bibinfo{title}{Achieving impact from ecosystem assessment and valuation of urban greenspace: The case of i-{T}ree {E}co in {G}reat {B}ritain}.
\newblock \bibinfo{journal}{Landscape and Urban Planning} \bibinfo{volume}{190}, \bibinfo{pages}{103590}.
%Type = Article
\bibitem[{Rauschnabel et~al.(2022)Rauschnabel, Felix, Hinsch, Shahab and Alt}]{rauschnabel2022xr}
\bibinfo{author}{Rauschnabel, P.A.}, \bibinfo{author}{Felix, R.}, \bibinfo{author}{Hinsch, C.}, \bibinfo{author}{Shahab, H.}, \bibinfo{author}{Alt, F.}, \bibinfo{year}{2022}.
\newblock \bibinfo{title}{What is {XR}? {T}owards a framework for augmented and virtual reality}.
\newblock \bibinfo{journal}{Computers in Human Behavior} \bibinfo{volume}{133}, \bibinfo{pages}{107289}.
%Type = Inproceedings
\bibitem[{Rish et~al.(2001)}]{rish2001empirical}
\bibinfo{author}{Rish, I.}, et~al., \bibinfo{year}{2001}.
\newblock \bibinfo{title}{An empirical study of the naive bayes classifier}, in: \bibinfo{booktitle}{Workshop on Empirical Methods in Artificial Intelligence}, \bibinfo{organization}{Citeseer}. pp. \bibinfo{pages}{41--46}.
%Type = Article
\bibitem[{Robertson(2004)}]{robertson2004understanding}
\bibinfo{author}{Robertson, S.}, \bibinfo{year}{2004}.
\newblock \bibinfo{title}{Understanding inverse document frequency: on theoretical arguments for {IDF}}.
\newblock \bibinfo{journal}{Journal of Documentation} \bibinfo{volume}{60}, \bibinfo{pages}{503--520}.
%Type = Article
\bibitem[{Rodriguez-Galiano et~al.(2012)Rodriguez-Galiano, Ghimire, Rogan, Chica-Olmo and Rigol-Sanchez}]{rodriguez2012assessment}
\bibinfo{author}{Rodriguez-Galiano, V.F.}, \bibinfo{author}{Ghimire, B.}, \bibinfo{author}{Rogan, J.}, \bibinfo{author}{Chica-Olmo, M.}, \bibinfo{author}{Rigol-Sanchez, J.P.}, \bibinfo{year}{2012}.
\newblock \bibinfo{title}{An assessment of the effectiveness of a random forest classifier for land-cover classification}.
\newblock \bibinfo{journal}{Journal of Photogrammetry and Remote Sensing} \bibinfo{volume}{67}, \bibinfo{pages}{93--104}.
%Type = Article
\bibitem[{Roh et~al.(2021)Roh, Heo and Whang}]{Roh202survey}
\bibinfo{author}{Roh, Y.}, \bibinfo{author}{Heo, G.}, \bibinfo{author}{Whang, S.E.}, \bibinfo{year}{2021}.
\newblock \bibinfo{title}{A survey on data collection for machine learning: A big data - ai integration perspective}.
\newblock \bibinfo{journal}{IEEE Transactions on Knowledge and Data Engineering} \bibinfo{volume}{33}, \bibinfo{pages}{1328--1347}.
%Type = Article
\bibitem[{Roman et~al.(2020)Roman, Bre, Fachinotti and Lamberts}]{roman2020application}
\bibinfo{author}{Roman, N.D.}, \bibinfo{author}{Bre, F.}, \bibinfo{author}{Fachinotti, V.D.}, \bibinfo{author}{Lamberts, R.}, \bibinfo{year}{2020}.
\newblock \bibinfo{title}{Application and characterization of metamodels based on artificial neural networks for building performance simulation: A systematic review}.
\newblock \bibinfo{journal}{Energy and Buildings} \bibinfo{volume}{217}, \bibinfo{pages}{109972}.
%Type = Book
\bibitem[{Van~der Ryn and Cowan(2013)}]{van2013ecological}
\bibinfo{author}{Van~der Ryn, S.}, \bibinfo{author}{Cowan, S.}, \bibinfo{year}{2013}.
\newblock \bibinfo{title}{Ecological design}.
\newblock \bibinfo{publisher}{Island press}.
%Type = Article
\bibitem[{Ryzhakova et~al.(2022)Ryzhakova, Malykhina, Pokolenko, Nesterenko and Honcharenko}]{ryzhakova2022construction}
\bibinfo{author}{Ryzhakova, G.}, \bibinfo{author}{Malykhina, O.}, \bibinfo{author}{Pokolenko, V.}, \bibinfo{author}{Nesterenko, I.}, \bibinfo{author}{Honcharenko, T.}, \bibinfo{year}{2022}.
\newblock \bibinfo{title}{Construction project management with digital twin information system}.
\newblock \bibinfo{journal}{International Journal of Emerging Technology and Advanced Engineering} \bibinfo{volume}{12}, \bibinfo{pages}{19--28}.
%Type = Article
\bibitem[{Salomon et~al.(1991)Salomon, Perkins and Globerson}]{salomon1991partners}
\bibinfo{author}{Salomon, G.}, \bibinfo{author}{Perkins, D.N.}, \bibinfo{author}{Globerson, T.}, \bibinfo{year}{1991}.
\newblock \bibinfo{title}{Partners in cognition: Extending human intelligence with intelligent technologies}.
\newblock \bibinfo{journal}{Educational researcher} \bibinfo{volume}{20}, \bibinfo{pages}{2--9}.
%Type = Article
\bibitem[{Schiefer et~al.(2020)Schiefer, Kattenborn, Frick, Frey, Schall, Koch and Schmidtlein}]{schiefer2020mapping}
\bibinfo{author}{Schiefer, F.}, \bibinfo{author}{Kattenborn, T.}, \bibinfo{author}{Frick, A.}, \bibinfo{author}{Frey, J.}, \bibinfo{author}{Schall, P.}, \bibinfo{author}{Koch, B.}, \bibinfo{author}{Schmidtlein, S.}, \bibinfo{year}{2020}.
\newblock \bibinfo{title}{Mapping forest tree species in high resolution {UAV}-based {RGB}-imagery by means of convolutional neural networks}.
\newblock \bibinfo{journal}{ISPRS Journal of Photogrammetry and Remote Sensing} \bibinfo{volume}{170}, \bibinfo{pages}{205--215}.
%Type = Article
\bibitem[{Scholz and Smith(2016)}]{scholz2016augmented}
\bibinfo{author}{Scholz, J.}, \bibinfo{author}{Smith, A.N.}, \bibinfo{year}{2016}.
\newblock \bibinfo{title}{Augmented reality: Designing immersive experiences that maximize consumer engagement}.
\newblock \bibinfo{journal}{Business Horizons} \bibinfo{volume}{59}, \bibinfo{pages}{149--161}.
%Type = Article
\bibitem[{Schuster and Paliwal(1997)}]{schuster1997bidirectional}
\bibinfo{author}{Schuster, M.}, \bibinfo{author}{Paliwal, K.K.}, \bibinfo{year}{1997}.
\newblock \bibinfo{title}{Bidirectional recurrent neural networks}.
\newblock \bibinfo{journal}{IEEE Transactions on Signal Processing} \bibinfo{volume}{45}, \bibinfo{pages}{2673--2681}.
%Type = Article
\bibitem[{Shi et~al.(2023)Shi, Shang and Qi}]{shi2023Iintelligent}
\bibinfo{author}{Shi, Y.}, \bibinfo{author}{Shang, M.}, \bibinfo{author}{Qi, Z.}, \bibinfo{year}{2023}.
\newblock \bibinfo{title}{Intelligent layout generation based on deep generative models: A comprehensive survey}.
\newblock \bibinfo{journal}{Information Fusion} \bibinfo{volume}{100}, \bibinfo{pages}{101940}.
%Type = Article
\bibitem[{Son et~al.(2023)Son, Weedon, Yigitcanlar, Sanchez, Corchado and Mehmood}]{son2023algorithmic}
\bibinfo{author}{Son, T.H.}, \bibinfo{author}{Weedon, Z.}, \bibinfo{author}{Yigitcanlar, T.}, \bibinfo{author}{Sanchez, T.}, \bibinfo{author}{Corchado, J.M.}, \bibinfo{author}{Mehmood, R.}, \bibinfo{year}{2023}.
\newblock \bibinfo{title}{Algorithmic urban planning for smart and sustainable development: Systematic review of the literature}.
\newblock \bibinfo{journal}{Sustainable Cities and Society} \bibinfo{volume}{94}, \bibinfo{pages}{104562}.
%Type = Article
\bibitem[{Sporleder(2010)}]{sporleder2010natural}
\bibinfo{author}{Sporleder, C.}, \bibinfo{year}{2010}.
\newblock \bibinfo{title}{Natural language processing for cultural heritage domains}.
\newblock \bibinfo{journal}{Language and Linguistics Compass} \bibinfo{volume}{4}, \bibinfo{pages}{750--768}.
%Type = Article
\bibitem[{Stupariu et~al.(2022)Stupariu, Cushman, Ple{\c{s}}oianu, P{\u{a}}tru~Stupariu and Fuerst}]{stupariu2022machine}
\bibinfo{author}{Stupariu, M.S.}, \bibinfo{author}{Cushman, S.A.}, \bibinfo{author}{Ple{\c{s}}oianu, A.I.}, \bibinfo{author}{P{\u{a}}tru~Stupariu, I.}, \bibinfo{author}{Fuerst, C.}, \bibinfo{year}{2022}.
\newblock \bibinfo{title}{Machine learning in landscape ecological analysis: a review of recent approaches}.
\newblock \bibinfo{journal}{Landscape Ecology} \bibinfo{volume}{37}, \bibinfo{pages}{1227--1250}.
%Type = Article
\bibitem[{Sun et~al.(2022)Sun, Gan, Chao and Yu}]{sun2022metaverse}
\bibinfo{author}{Sun, J.}, \bibinfo{author}{Gan, W.}, \bibinfo{author}{Chao, H.C.}, \bibinfo{author}{Yu, P.S.}, \bibinfo{year}{2022}.
\newblock \bibinfo{title}{Metaverse: Survey, applications, security, and opportunities}.
\newblock \bibinfo{journal}{arXiv preprint arXiv:2210.07990} .
%Type = Article
\bibitem[{Sun et~al.(2023)Sun, Gan, Chao, Yu and Ding}]{sun2023internet}
\bibinfo{author}{Sun, J.}, \bibinfo{author}{Gan, W.}, \bibinfo{author}{Chao, H.C.}, \bibinfo{author}{Yu, P.S.}, \bibinfo{author}{Ding, W.}, \bibinfo{year}{2023}.
\newblock \bibinfo{title}{Internet of behaviors: A survey}.
\newblock \bibinfo{journal}{IEEE Internet of Things Journal} \bibinfo{volume}{10}, \bibinfo{pages}{11117--11134}.
%Type = Book
\bibitem[{Sutton and Barto(2018)}]{sutton2018reinforcement}
\bibinfo{author}{Sutton, R.S.}, \bibinfo{author}{Barto, A.G.}, \bibinfo{year}{2018}.
\newblock \bibinfo{title}{Reinforcement learning: {A}n introduction}.
\newblock \bibinfo{publisher}{MIT press}.
%Type = Inproceedings
\bibitem[{Szegedy et~al.(2016)Szegedy, Vanhoucke, Ioffe, Shlens and Wojna}]{szegedy2016rethinking}
\bibinfo{author}{Szegedy, C.}, \bibinfo{author}{Vanhoucke, V.}, \bibinfo{author}{Ioffe, S.}, \bibinfo{author}{Shlens, J.}, \bibinfo{author}{Wojna, Z.}, \bibinfo{year}{2016}.
\newblock \bibinfo{title}{Rethinking the inception architecture for computer vision}, in: \bibinfo{booktitle}{The IEEE Conference on Computer Vision and Pattern Recognition}, pp. \bibinfo{pages}{2818--2826}.
%Type = Book
\bibitem[{Szeliski(2022)}]{szeliski2022computer}
\bibinfo{author}{Szeliski, R.}, \bibinfo{year}{2022}.
\newblock \bibinfo{title}{Computer vision: algorithms and applications}.
\newblock \bibinfo{publisher}{Springer Nature}.
%Type = Article
\bibitem[{Talaviya et~al.(2020)Talaviya, Shah, Patel, Yagnik and Shah}]{talaviya2020implementation}
\bibinfo{author}{Talaviya, T.}, \bibinfo{author}{Shah, D.}, \bibinfo{author}{Patel, N.}, \bibinfo{author}{Yagnik, H.}, \bibinfo{author}{Shah, M.}, \bibinfo{year}{2020}.
\newblock \bibinfo{title}{Implementation of artificial intelligence in agriculture for optimisation of irrigation and application of pesticides and herbicides}.
\newblock \bibinfo{journal}{Artificial Intelligence in Agriculture} \bibinfo{volume}{4}, \bibinfo{pages}{58--73}.
%Type = Article
\bibitem[{Talib et~al.(2021)Talib, Majzoub, Nasir and Jamal}]{talib2021systematic}
\bibinfo{author}{Talib, M.A.}, \bibinfo{author}{Majzoub, S.}, \bibinfo{author}{Nasir, Q.}, \bibinfo{author}{Jamal, D.}, \bibinfo{year}{2021}.
\newblock \bibinfo{title}{A systematic literature review on hardware implementation of artificial intelligence algorithms}.
\newblock \bibinfo{journal}{The Journal of Supercomputing} \bibinfo{volume}{77}, \bibinfo{pages}{1897--1938}.
%Type = Article
\bibitem[{Tan and Cheng(2024)}]{tan2024digital}
\bibinfo{author}{Tan, F.}, \bibinfo{author}{Cheng, Y.}, \bibinfo{year}{2024}.
\newblock \bibinfo{title}{A digital twin framework for innovating rural ecological landscape control}.
\newblock \bibinfo{journal}{Environmental Sciences Europe} \bibinfo{volume}{36}, \bibinfo{pages}{59}.
%Type = Inproceedings
\bibitem[{Tan and Cao(2021)}]{tan2021efficient}
\bibinfo{author}{Tan, T.}, \bibinfo{author}{Cao, G.}, \bibinfo{year}{2021}.
\newblock \bibinfo{title}{Efficient execution of deep neural networks on mobile devices with {NPU}}, in: \bibinfo{booktitle}{the 20th International Conference on Information Processing in Sensor Networks}, pp. \bibinfo{pages}{283--298}.
%Type = Article
\bibitem[{Taylor and Lennon(2011)}]{taylor2011cultural}
\bibinfo{author}{Taylor, K.}, \bibinfo{author}{Lennon, J.}, \bibinfo{year}{2011}.
\newblock \bibinfo{title}{Cultural landscapes: a bridge between culture and nature?}
\newblock \bibinfo{journal}{International Journal of Heritage Studies} \bibinfo{volume}{17}, \bibinfo{pages}{537--554}.
%Type = Article
\bibitem[{Tejeda et~al.(2022)Tejeda, Kumar, Smyth and Steyvers}]{tejeda2022ai}
\bibinfo{author}{Tejeda, H.}, \bibinfo{author}{Kumar, A.}, \bibinfo{author}{Smyth, P.}, \bibinfo{author}{Steyvers, M.}, \bibinfo{year}{2022}.
\newblock \bibinfo{title}{{AI}-assisted decision-making: A cognitive modeling approach to infer latent reliance strategies}.
\newblock \bibinfo{journal}{Computational Brain \& Behavior} \bibinfo{volume}{5}, \bibinfo{pages}{491--508}.
%Type = Article
\bibitem[{Tong et~al.(2020)Tong, Xia, Lu, Shen, Li, You and Zhang}]{tong2020land}
\bibinfo{author}{Tong, X.}, \bibinfo{author}{Xia, G.}, \bibinfo{author}{Lu, Q.}, \bibinfo{author}{Shen, H.}, \bibinfo{author}{Li, S.}, \bibinfo{author}{You, S.}, \bibinfo{author}{Zhang, L.}, \bibinfo{year}{2020}.
\newblock \bibinfo{title}{Land-cover classification with high-resolution remote sensing images using transferable deep models}.
\newblock \bibinfo{journal}{Remote Sensing of Environment} \bibinfo{volume}{237}, \bibinfo{pages}{111322}.
%Type = Article
\bibitem[{Turner(1989)}]{turner1989landscape}
\bibinfo{author}{Turner, M.G.}, \bibinfo{year}{1989}.
\newblock \bibinfo{title}{Landscape ecology: the effect of pattern on process}.
\newblock \bibinfo{journal}{Annual Review of Ecology and Systematics} \bibinfo{volume}{20}, \bibinfo{pages}{171--197}.
%Type = Article
\bibitem[{Ullo and Sinha(2020)}]{ullo2020advances}
\bibinfo{author}{Ullo, S.L.}, \bibinfo{author}{Sinha, G.R.}, \bibinfo{year}{2020}.
\newblock \bibinfo{title}{Advances in smart environment monitoring systems using {I}o{T} and sensors}.
\newblock \bibinfo{journal}{Sensors} \bibinfo{volume}{20}, \bibinfo{pages}{3113}.
%Type = Inproceedings
\bibitem[{Urban~Davis et~al.(2021)Urban~Davis, Anderson, Stroetzel, Grossman and Fitzmaurice}]{urban2021designing}
\bibinfo{author}{Urban~Davis, J.}, \bibinfo{author}{Anderson, F.}, \bibinfo{author}{Stroetzel, M.}, \bibinfo{author}{Grossman, T.}, \bibinfo{author}{Fitzmaurice, G.}, \bibinfo{year}{2021}.
\newblock \bibinfo{title}{Designing co-creative ai for virtual environments}, in: \bibinfo{booktitle}{The 13th Conference on Creativity and Cognition}, pp. \bibinfo{pages}{1--11}.
%Type = Article
\bibitem[{Vansteenwegen et~al.(2011)Vansteenwegen, Souffriau, Berghe and Van~Oudheusden}]{vansteenwegen2011city}
\bibinfo{author}{Vansteenwegen, P.}, \bibinfo{author}{Souffriau, W.}, \bibinfo{author}{Berghe, G.V.}, \bibinfo{author}{Van~Oudheusden, D.}, \bibinfo{year}{2011}.
\newblock \bibinfo{title}{The city trip planner: an expert system for tourists}.
\newblock \bibinfo{journal}{Expert Systems with Applications} \bibinfo{volume}{38}, \bibinfo{pages}{6540--6546}.
%Type = Article
\bibitem[{Volk et~al.(2014)Volk, Stengel and Schultmann}]{volk2014building}
\bibinfo{author}{Volk, R.}, \bibinfo{author}{Stengel, J.}, \bibinfo{author}{Schultmann, F.}, \bibinfo{year}{2014}.
\newblock \bibinfo{title}{Building information modeling ({BIM}) for existing buildings—literature review and future needs}.
\newblock \bibinfo{journal}{Automation in Construction} \bibinfo{volume}{38}, \bibinfo{pages}{109--127}.
%Type = Article
\bibitem[{Wang(2022)}]{wang2022artificial}
\bibinfo{author}{Wang, X.}, \bibinfo{year}{2022}.
\newblock \bibinfo{title}{Artificial intelligence in the protection and inheritance of cultural landscape heritage in traditional village}.
\newblock \bibinfo{journal}{Scientific Programming} \bibinfo{volume}{2022}, \bibinfo{pages}{9117981}.
%Type = Article
\bibitem[{Wang et~al.(2013)Wang, Love, Kim, Park, Sing and Hou}]{wang2013conceptual}
\bibinfo{author}{Wang, X.}, \bibinfo{author}{Love, P.E.}, \bibinfo{author}{Kim, M.J.}, \bibinfo{author}{Park, C.S.}, \bibinfo{author}{Sing, C.P.}, \bibinfo{author}{Hou, L.}, \bibinfo{year}{2013}.
\newblock \bibinfo{title}{A conceptual framework for integrating building information modeling with augmented reality}.
\newblock \bibinfo{journal}{Automation in Cconstruction} \bibinfo{volume}{34}, \bibinfo{pages}{37--44}.
%Type = Inproceedings
\bibitem[{Wang(2017)}]{wang2017application}
\bibinfo{author}{Wang, Y.}, \bibinfo{year}{2017}.
\newblock \bibinfo{title}{Application research on urban cultural landscape heritage protection using digital technology}, in: \bibinfo{booktitle}{International Conference on Robots \& Intelligent System}, \bibinfo{organization}{IEEE}. pp. \bibinfo{pages}{58--61}.
%Type = Article
\bibitem[{Wang et~al.(2022)Wang, Su, Zhang, Xing, Liu, Luan and Shen}]{wang2022survey}
\bibinfo{author}{Wang, Y.}, \bibinfo{author}{Su, Z.}, \bibinfo{author}{Zhang, N.}, \bibinfo{author}{Xing, R.}, \bibinfo{author}{Liu, D.}, \bibinfo{author}{Luan, T.H.}, \bibinfo{author}{Shen, X.}, \bibinfo{year}{2022}.
\newblock \bibinfo{title}{A survey on metaverse: Fundamentals, security, and privacy}.
\newblock \bibinfo{journal}{IEEE Communications Surveys \& Tutorials} \bibinfo{volume}{25}, \bibinfo{pages}{319--352}.
%Type = Inproceedings
\bibitem[{Wang et~al.(2020)Wang, Wang, Shi, He, Tang, Zhao and Chu}]{wang2020benchmarking}
\bibinfo{author}{Wang, Y.}, \bibinfo{author}{Wang, Q.}, \bibinfo{author}{Shi, S.}, \bibinfo{author}{He, X.}, \bibinfo{author}{Tang, Z.}, \bibinfo{author}{Zhao, K.}, \bibinfo{author}{Chu, X.}, \bibinfo{year}{2020}.
\newblock \bibinfo{title}{Benchmarking the performance and energy efficiency of {AI} accelerators for {AI} training}, in: \bibinfo{booktitle}{the 20th IEEE/ACM International Symposium on Cluster, Cloud and Internet Computing}, \bibinfo{organization}{IEEE}. pp. \bibinfo{pages}{744--751}.
%Type = Article
\bibitem[{Westphal et~al.(2007)Westphal, Field and Possingham}]{westphal2007optimizing}
\bibinfo{author}{Westphal, M.I.}, \bibinfo{author}{Field, S.A.}, \bibinfo{author}{Possingham, H.P.}, \bibinfo{year}{2007}.
\newblock \bibinfo{title}{Optimizing landscape configuration: a case study of woodland birds in the mount lofty ranges, south australia}.
\newblock \bibinfo{journal}{Landscape and Urban Planning} \bibinfo{volume}{81}, \bibinfo{pages}{56--66}.
%Type = Article
\bibitem[{Wik et~al.(2018)Wik, Sekse, Enebo and Thorvaldsen}]{wik2018bim}
\bibinfo{author}{Wik, K.H.}, \bibinfo{author}{Sekse, M.}, \bibinfo{author}{Enebo, B.A.}, \bibinfo{author}{Thorvaldsen, J.}, \bibinfo{year}{2018}.
\newblock \bibinfo{title}{Bim for landscape: a norwegian standardization project}.
\newblock \bibinfo{journal}{Journal of Digital Landscape Architecture} \bibinfo{volume}{3}, \bibinfo{pages}{241--248}.
%Type = Article
\bibitem[{Wolfert et~al.(2017)Wolfert, Ge, Verdouw and Bogaardt}]{wolfert2017big}
\bibinfo{author}{Wolfert, S.}, \bibinfo{author}{Ge, L.}, \bibinfo{author}{Verdouw, C.}, \bibinfo{author}{Bogaardt, M.J.}, \bibinfo{year}{2017}.
\newblock \bibinfo{title}{Big data in smart farming--a review}.
\newblock \bibinfo{journal}{Agricultural Systems} \bibinfo{volume}{153}, \bibinfo{pages}{69--80}.
%Type = Article
\bibitem[{Wolfram(1984)}]{wolfram1984cellular}
\bibinfo{author}{Wolfram, S.}, \bibinfo{year}{1984}.
\newblock \bibinfo{title}{Cellular automata as models of complexity}.
\newblock \bibinfo{journal}{Nature} \bibinfo{volume}{311}, \bibinfo{pages}{419--424}.
%Type = Book
\bibitem[{Wong and Lee(2005)}]{wong2005statistical}
\bibinfo{author}{Wong, W.}, \bibinfo{author}{Lee, J.}, \bibinfo{year}{2005}.
\newblock \bibinfo{title}{Statistical analysis of geographic information with ArcView {GIS} and ArcGIS}.
\newblock \bibinfo{publisher}{Wiley}.
%Type = Article
\bibitem[{Wu(2013)}]{wu2013landscape}
\bibinfo{author}{Wu, J.}, \bibinfo{year}{2013}.
\newblock \bibinfo{title}{Landscape sustainability science: ecosystem services and human well-being in changing landscapes}.
\newblock \bibinfo{journal}{Landscape Ecology} \bibinfo{volume}{28}, \bibinfo{pages}{999--1023}.
%Type = Article
\bibitem[{Wu et~al.(2024)Wu, Gan, Chao and Yu}]{wu2024geospatial}
\bibinfo{author}{Wu, J.}, \bibinfo{author}{Gan, W.}, \bibinfo{author}{Chao, H.C.}, \bibinfo{author}{Yu, P.S.}, \bibinfo{year}{2024}.
\newblock \bibinfo{title}{Geospatial big data: Survey and challenges}.
\newblock \bibinfo{journal}{IEEE Journal of Selected Topics in Applied Earth Observations and Remote Sensing} , \bibinfo{pages}{1--14}.
%Type = Article
\bibitem[{Wu et~al.(2023)Wu, Gan, Chen, Wan and Lin}]{wu2023ai}
\bibinfo{author}{Wu, J.}, \bibinfo{author}{Gan, W.}, \bibinfo{author}{Chen, Z.}, \bibinfo{author}{Wan, S.}, \bibinfo{author}{Lin, H.}, \bibinfo{year}{2023}.
\newblock \bibinfo{title}{{AI}-generated content ({AIGC}): A survey}.
\newblock \bibinfo{journal}{arXiv preprint arXiv:2304.06632} .
%Type = Article
\bibitem[{Wu and Shang(2020)}]{wu2020managing}
\bibinfo{author}{Wu, J.}, \bibinfo{author}{Shang, S.}, \bibinfo{year}{2020}.
\newblock \bibinfo{title}{Managing uncertainty in {AI}-enabled decision making and achieving sustainability}.
\newblock \bibinfo{journal}{Sustainability} \bibinfo{volume}{12}, \bibinfo{pages}{8758}.
%Type = Article
\bibitem[{Wu et~al.(2019)Wu, Wang, Qiu and Peng}]{wu2019using}
\bibinfo{author}{Wu, J.}, \bibinfo{author}{Wang, Y.}, \bibinfo{author}{Qiu, S.}, \bibinfo{author}{Peng, J.}, \bibinfo{year}{2019}.
\newblock \bibinfo{title}{Using the modified i-{T}ree {E}co model to quantify air pollution removal by urban vegetation}.
\newblock \bibinfo{journal}{Science of the Total Environment} \bibinfo{volume}{688}, \bibinfo{pages}{673--683}.
%Type = Article
\bibitem[{Wu et~al.(2013)Wu, Zhu, Wu and Ding}]{wu2013data}
\bibinfo{author}{Wu, X.}, \bibinfo{author}{Zhu, X.}, \bibinfo{author}{Wu, G.Q.}, \bibinfo{author}{Ding, W.}, \bibinfo{year}{2013}.
\newblock \bibinfo{title}{Data mining with big data}.
\newblock \bibinfo{journal}{IEEE Transactions on Knowledge and Data Engineering} \bibinfo{volume}{26}, \bibinfo{pages}{97--107}.
%Type = Article
\bibitem[{Xia et~al.(2021)Xia, Yabuki and Fukuda}]{xia2021development}
\bibinfo{author}{Xia, Y.}, \bibinfo{author}{Yabuki, N.}, \bibinfo{author}{Fukuda, T.}, \bibinfo{year}{2021}.
\newblock \bibinfo{title}{Development of a system for assessing the quality of urban street-level greenery using street view images and deep learning}.
\newblock \bibinfo{journal}{Urban Forestry \& Urban Greening} \bibinfo{volume}{59}, \bibinfo{pages}{126995}.
%Type = Article
\bibitem[{Xu et~al.(2021)Xu, Liu, Cao, Huang, Liu, Qian, Liu, Wu, Dong, Qiu et~al.}]{xu2021artificial}
\bibinfo{author}{Xu, Y.}, \bibinfo{author}{Liu, X.}, \bibinfo{author}{Cao, X.}, \bibinfo{author}{Huang, C.}, \bibinfo{author}{Liu, E.}, \bibinfo{author}{Qian, S.}, \bibinfo{author}{Liu, X.}, \bibinfo{author}{Wu, Y.}, \bibinfo{author}{Dong, F.}, \bibinfo{author}{Qiu, C.W.}, et~al., \bibinfo{year}{2021}.
\newblock \bibinfo{title}{Artificial intelligence: A powerful paradigm for scientific research}.
\newblock \bibinfo{journal}{The Innovation} \bibinfo{volume}{2}, \bibinfo{pages}{100179}.
%Type = Article
\bibitem[{Yang et~al.(2020)Yang, Gao, Li and Van~Eetvelde}]{yang2020multi}
\bibinfo{author}{Yang, D.}, \bibinfo{author}{Gao, C.}, \bibinfo{author}{Li, L.}, \bibinfo{author}{Van~Eetvelde, V.}, \bibinfo{year}{2020}.
\newblock \bibinfo{title}{Multi-scaled identification of landscape character types and areas in {L}ushan {N}ational {P}ark and its fringes, {C}hina}.
\newblock \bibinfo{journal}{Landscape and Urban Planning} \bibinfo{volume}{201}, \bibinfo{pages}{103844}.
%Type = Article
\bibitem[{Yao et~al.(2017)Yao, Li, Liu, Liu, Liang, Zhang and Mai}]{yao2017sensing}
\bibinfo{author}{Yao, Y.}, \bibinfo{author}{Li, X.}, \bibinfo{author}{Liu, X.}, \bibinfo{author}{Liu, P.}, \bibinfo{author}{Liang, Z.}, \bibinfo{author}{Zhang, J.}, \bibinfo{author}{Mai, K.}, \bibinfo{year}{2017}.
\newblock \bibinfo{title}{Sensing spatial distribution of urban land use by integrating points-of-interest and google {W}ord2{V}ec model}.
\newblock \bibinfo{journal}{International Journal of Geographical Information Science} \bibinfo{volume}{31}, \bibinfo{pages}{825--848}.
%Type = Article
\bibitem[{Ye et~al.(2023)Ye, Du, Han, Newman, Retchless, Zou, Ham and Cai}]{ye2023developing}
\bibinfo{author}{Ye, X.}, \bibinfo{author}{Du, J.}, \bibinfo{author}{Han, Y.}, \bibinfo{author}{Newman, G.}, \bibinfo{author}{Retchless, D.}, \bibinfo{author}{Zou, L.}, \bibinfo{author}{Ham, Y.}, \bibinfo{author}{Cai, Z.}, \bibinfo{year}{2023}.
\newblock \bibinfo{title}{Developing human-centered urban digital twins for community infrastructure resilience: A research agenda}.
\newblock \bibinfo{journal}{Journal of Planning Literature} \bibinfo{volume}{38}, \bibinfo{pages}{187--199}.
%Type = Article
\bibitem[{Ye et~al.(2022)Ye, Du and Ye}]{ye2022masterplangan}
\bibinfo{author}{Ye, X.}, \bibinfo{author}{Du, J.}, \bibinfo{author}{Ye, Y.}, \bibinfo{year}{2022}.
\newblock \bibinfo{title}{Masterplan{GAN}: Facilitating the smart rendering of urban master plans via generative adversarial networks}.
\newblock \bibinfo{journal}{Environment and Planning B: Urban Analytics and City Science} \bibinfo{volume}{49}, \bibinfo{pages}{794--814}.
%Type = Article
\bibitem[{Zhai et~al.(2020)Zhai, Yao, Guan, Liang, Li, Pan, Yue, Yuan and Zhou}]{zhai2020simulating}
\bibinfo{author}{Zhai, Y.}, \bibinfo{author}{Yao, Y.}, \bibinfo{author}{Guan, Q.}, \bibinfo{author}{Liang, X.}, \bibinfo{author}{Li, X.}, \bibinfo{author}{Pan, Y.}, \bibinfo{author}{Yue, H.}, \bibinfo{author}{Yuan, Z.}, \bibinfo{author}{Zhou, J.}, \bibinfo{year}{2020}.
\newblock \bibinfo{title}{Simulating urban land use change by integrating a convolutional neural network with vector-based cellular automata}.
\newblock \bibinfo{journal}{International Journal of Geographical Information Science} \bibinfo{volume}{34}, \bibinfo{pages}{1475--1499}.
%Type = Article
\bibitem[{Zhang and Lu(2021)}]{zhang2021study}
\bibinfo{author}{Zhang, C.}, \bibinfo{author}{Lu, Y.}, \bibinfo{year}{2021}.
\newblock \bibinfo{title}{Study on artificial intelligence: The state of the art and future prospects}.
\newblock \bibinfo{journal}{Journal of Industrial Information Integration} \bibinfo{volume}{23}, \bibinfo{pages}{100224}.
%Type = Article
\bibitem[{Zhang et~al.(2019a)Zhang, Sargent, Pan, Li, Gardiner, Hare and Atkinson}]{zhang2019joint}
\bibinfo{author}{Zhang, C.}, \bibinfo{author}{Sargent, I.}, \bibinfo{author}{Pan, X.}, \bibinfo{author}{Li, H.}, \bibinfo{author}{Gardiner, A.}, \bibinfo{author}{Hare, J.}, \bibinfo{author}{Atkinson, P.M.}, \bibinfo{year}{2019}a.
\newblock \bibinfo{title}{Joint deep learning for land cover and land use classification}.
\newblock \bibinfo{journal}{Remote Sensing of Environment} \bibinfo{volume}{221}, \bibinfo{pages}{173--187}.
%Type = Article
\bibitem[{Zhang et~al.(2018)Zhang, Zhou, Liu, Liu, Fung, Lin and Ratti}]{zhang2018measuring}
\bibinfo{author}{Zhang, F.}, \bibinfo{author}{Zhou, B.}, \bibinfo{author}{Liu, L.}, \bibinfo{author}{Liu, Y.}, \bibinfo{author}{Fung, H.H.}, \bibinfo{author}{Lin, H.}, \bibinfo{author}{Ratti, C.}, \bibinfo{year}{2018}.
\newblock \bibinfo{title}{Measuring human perceptions of a large-scale urban region using machine learning}.
\newblock \bibinfo{journal}{Landscape and Urban Planning} \bibinfo{volume}{180}, \bibinfo{pages}{148--160}.
%Type = Article
\bibitem[{Zhang and Aslan(2021)}]{zhang2021ai}
\bibinfo{author}{Zhang, K.}, \bibinfo{author}{Aslan, A.B.}, \bibinfo{year}{2021}.
\newblock \bibinfo{title}{{AI} technologies for education: Recent research \& future directions}.
\newblock \bibinfo{journal}{Computers and Education: Artificial Intelligence} \bibinfo{volume}{2}, \bibinfo{pages}{100025}.
%Type = Article
\bibitem[{Zhang et~al.(2019b)Zhang, Zhang, Jeng and Zeng}]{zhang2019cityscape}
\bibinfo{author}{Zhang, L.M.}, \bibinfo{author}{Zhang, R.X.}, \bibinfo{author}{Jeng, T.S.}, \bibinfo{author}{Zeng, Z.Y.}, \bibinfo{year}{2019}b.
\newblock \bibinfo{title}{Cityscape protection using {VR} and eye tracking technology}.
\newblock \bibinfo{journal}{Journal of Visual Communication and Image Representation} \bibinfo{volume}{64}, \bibinfo{pages}{102639}.
%Type = Article
\bibitem[{Zhao et~al.(2022)Zhao, Li, Sun, Zhao, Gai, Wang, Huang, Yu, Wang, Zhang et~al.}]{zhao2022intelligent}
\bibinfo{author}{Zhao, X.}, \bibinfo{author}{Li, M.}, \bibinfo{author}{Sun, Z.}, \bibinfo{author}{Zhao, Y.}, \bibinfo{author}{Gai, Y.}, \bibinfo{author}{Wang, J.}, \bibinfo{author}{Huang, C.}, \bibinfo{author}{Yu, L.}, \bibinfo{author}{Wang, S.}, \bibinfo{author}{Zhang, M.}, et~al., \bibinfo{year}{2022}.
\newblock \bibinfo{title}{Intelligent construction and management of landscapes through building information modeling and mixed reality}.
\newblock \bibinfo{journal}{Applied Sciences} \bibinfo{volume}{12}, \bibinfo{pages}{7118}.
%Type = Article
\bibitem[{Zheng and Yang(2020)}]{Zheng2020Research_CN}
\bibinfo{author}{Zheng, Y.}, \bibinfo{author}{Yang, J.}, \bibinfo{year}{2020}.
\newblock \bibinfo{title}{Research on elaborate urban repair planning approach based on ai analysis of large-scale street-view big data}.
\newblock \bibinfo{journal}{Chinese Landscape Architecture} \bibinfo{volume}{36}, \bibinfo{pages}{73}.
%Type = Article
\bibitem[{Zhong et~al.(2021)Zhong, Wang and Zhang}]{zhong2021application}
\bibinfo{author}{Zhong, H.}, \bibinfo{author}{Wang, L.}, \bibinfo{author}{Zhang, H.}, \bibinfo{year}{2021}.
\newblock \bibinfo{title}{The application of virtual reality technology in the digital preservation of cultural heritage}.
\newblock \bibinfo{journal}{Computer Science and Information Systems} \bibinfo{volume}{18}, \bibinfo{pages}{535--551}.
%Type = Article
\bibitem[{Zhu and Wu(2005)}]{zhu2005cost}
\bibinfo{author}{Zhu, X.}, \bibinfo{author}{Wu, X.}, \bibinfo{year}{2005}.
\newblock \bibinfo{title}{Cost-constrained data acquisition for intelligent data preparation}.
\newblock \bibinfo{journal}{IEEE Transactions on Knowledge and Data Engineering} \bibinfo{volume}{17}, \bibinfo{pages}{1542--1556}.
%Type = Article
\bibitem[{Zuo and Zhao(2014)}]{zuo2014green}
\bibinfo{author}{Zuo, J.}, \bibinfo{author}{Zhao, Z.}, \bibinfo{year}{2014}.
\newblock \bibinfo{title}{Green building research--current status and future agenda: A review}.
\newblock \bibinfo{journal}{Renewable and Sustainable Energy Reviews} \bibinfo{volume}{30}, \bibinfo{pages}{271--281}.

\end{thebibliography}

\end{document}